\pdfoutput=1
\documentclass[11pt]{article}
\usepackage[utf8]{inputenc}
\usepackage[T1]{fontenc}
\usepackage{lmodern}
\usepackage{authblk}
\usepackage{setspace}
\usepackage[margin=1in]{geometry}
\usepackage{graphicx}
\usepackage{amsmath,amssymb,amsthm}
\usepackage{bm}
\usepackage{booktabs}
\usepackage{array}
\usepackage{multirow}
\usepackage[section]{placeins}
\usepackage{algorithm}
\usepackage{algpseudocode}
\usepackage[labelfont=bf,labelsep=period,font=small]{caption}
\usepackage{xcolor}
\usepackage[round,authoryear]{natbib}
\usepackage[colorlinks=true,linkcolor=blue!50!black,citecolor=blue!50!black,urlcolor=blue!50!black]{hyperref}

\newcommand{\model}{MJBoost}
\newcommand{\E}{\mathbb{E}}
\newcommand{\Var}{\operatorname{Var}}
\newcommand{\ablBB}{75.7\%}
\newcommand{\ablHess}{4.1\%}
\newcommand{\ablMtwo}{15.6\%}
\newcommand{\ablPrune}{1.9\%}
\newcommand{\caseAboveNorth}{300}
\newcommand{\caseAboveSlope}{127}
\newcommand{\caseAboveSouth}{268}
\newcommand{\caseCheck}{4\times10^{-7}}
\newcommand{\caseD}{21}
\newcommand{\caseGradeSevenEleven}{360}
\newcommand{\caseLocHi}{+522}
\newcommand{\caseLocLo}{$-$248}
\newcommand{\caseN}{21{,}613}
\newcommand{\caseRMSE}{100.2}
\newcommand{\caseRMSEebm}{105.3}
\newcommand{\caseRMSElgbm}{100.6}
\newcommand{\caseRMSEmain}{125.1}
\newcommand{\caseRMSEtwo}{104.1}

\newcommand{\caseRtwo}{0.994}

\newcommand{\caseYrMid}{23}
\newcommand{\caseYrNew}{53}

\newcommand{\ccAucEBM}{0.895}
\newcommand{\ccAucKAU}{0.893}
\newcommand{\ccAucLGBM}{0.894}

\newcommand{\ccCpuEBM}{19.5}
\newcommand{\ccCpuKAU}{31.5}
\newcommand{\ccCpuLGBM}{3.3}

\newcommand{\ccDMax}{118}
\newcommand{\ccDMin}{4}
\newcommand{\ccFriedman}{0.15}

\newcommand{\ccLowerEBM}{17}

\newcommand{\ccMedRatioEBM}{$-$0.1\%}

\newcommand{\ccRankEBM}{2.16}
\newcommand{\ccRankKAU}{2.12}
\newcommand{\ccRankLGBM}{1.72}
\newcommand{\ccRowsMax}{96{,}320}
\newcommand{\ccRowsMin}{500}

\newcommand{\ccTasks}{32}

\newcommand{\cpuRatioEBM}{1.91}
\newcommand{\cpuRatioLGBM}{6.8}
\newcommand{\cpuTotalEBM}{68}
\newcommand{\cpuTotalKAU}{130}
\newcommand{\cpuTotalLGBM}{19}
\newcommand{\ctrAllFriedman}{6.4\times10^{-7}}
\newcommand{\ctrAllGmEBM}{$-$29.1\%}
\newcommand{\ctrAllGmEBMabs}{29.1\%}

\newcommand{\ctrAllGmLGBMabs}{13.2\%}
\newcommand{\ctrAllGmLGBMdir}{lower}
\newcommand{\ctrAllLossEBM}{2}
\newcommand{\ctrAllLossLGBM}{12}
\newcommand{\ctrAllRankEBM}{2.71}
\newcommand{\ctrAllRankKAU}{1.80}
\newcommand{\ctrAllRankLGBM}{1.49}
\newcommand{\ctrAllTasks}{35}
\newcommand{\ctrAllTiesEBM}{12}
\newcommand{\ctrAllTiesLGBM}{13}
\newcommand{\ctrAllWinsEBM}{21}
\newcommand{\ctrAllWinsLGBM}{10}
\newcommand{\ctrBeatLGBMlist}{socmob, Moneyball, auction verification, space\_ga, abalone, kin8nm, Brazilian houses, naval propulsion plant, FPS benchmark, SARCOS}

\newcommand{\ctrHeldGmEBMabs}{30.9\%}

\newcommand{\ctrHeldLossEBM}{2}

\newcommand{\ctrHeldTasks}{30}
\newcommand{\ctrHeldTiesEBM}{11}

\newcommand{\ctrHeldWinsEBM}{17}

\newcommand{\decBeatEBMfull}{30}
\newcommand{\decBeatEBMtwo}{16}
\newcommand{\decCheck}{4\times10^{-7}}
\newcommand{\decMainMed}{0.95}
\newcommand{\decPairMed}{0.057}
\newcommand{\decRestMed}{0.008}

\newcommand{\decRtwoMed}{0.992}
\newcommand{\decTasks}{35}
\newcommand{\decUpTwoLossMed}{4.1\%}
\newcommand{\decUpTwoTop}{FPS benchmark (+1981\%), video transcoding (+1166\%), auction verification (+401\%) and naval propulsion plant (+252\%)}

\newcommand{\figOneRMSE}{1.10}
\newcommand{\figOneRMSEebm}{1.55}
\newcommand{\figOneRMSEtwo}{3.41}

\newcommand{\figOneRtwo}{0.84}
\newcommand{\mcAccEBM}{0.782}
\newcommand{\mcAccKAU}{0.789}
\newcommand{\mcAccLGBM}{0.772}

\newcommand{\mcCpuEBM}{2.8}
\newcommand{\mcCpuKAU}{10.5}
\newcommand{\mcCpuLGBM}{1.2}
\newcommand{\mcDMax}{27}
\newcommand{\mcDMin}{4}
\newcommand{\mcFriedman}{0.23}
\newcommand{\mcGainBalance}{$-$73.0\%}
\newcommand{\mcGainCar}{$-$97.7\%}
\newcommand{\mcKMax}{11}
\newcommand{\mcKMin}{3}

\newcommand{\mcLossEBM}{3}
\newcommand{\mcLossLGBM}{1}
\newcommand{\mcLowerEBM}{8}
\newcommand{\mcLowerLGBM}{7}
\newcommand{\mcMedRatioEBM}{$-$5.1\%}

\newcommand{\mcMedRatioLGBM}{$-$7.8\%}

\newcommand{\mcRankEBM}{2.00}
\newcommand{\mcRankKAU}{1.64}
\newcommand{\mcRankLGBM}{2.36}
\newcommand{\mcRowsMax}{5{,}456}
\newcommand{\mcRowsMin}{625}
\newcommand{\mcTasks}{11}
\newcommand{\mcTiesEBM}{4}
\newcommand{\mcTiesLGBM}{3}
\newcommand{\mcWinsEBM}{4}
\newcommand{\mcWinsLGBM}{7}
\newcommand{\ordBestFold}{13}
\newcommand{\ordExpThreeEBM}{0.206}
\newcommand{\ordExpThreeKAU}{0.017}
\newcommand{\ordExpThreeUpTwo}{0.185}
\newcommand{\ordProdEBM}{1.00}
\newcommand{\ordProdFive}{0.56}
\newcommand{\ordProdSix}{0.82}

\newcommand{\ordShareErr}{0.04}
\newcommand{\ordSpurMax}{0.003}
\newcommand{\supMax}{12.0}
\newcommand{\supMin}{2.6}

\title{Gradient Boosting with Kolmogorov--Arnold Units\\for Interpretable Regression beyond Pairwise Interactions}
\author[1*]{Minjong Cheon}
\author[2]{Hanseon Joo}
\author[3]{Eunji Lee}
\affil[1]{Department of Computer Science, Korea National Open University, Seoul, South Korea}
\affil[2]{Department of Information Systems, Hanyang University, South Korea}
\affil[3]{Korea Racing Authority, South Korea}
\affil[*]{Correspondence: jmj2316@mail.knou.ac.kr}
\date{}

\begin{document}
\maketitle

\begin{abstract}
Interpretable models for tabular regression, such as the explainable boosting machine (EBM), restrict a
prediction to main effects and pairwise interactions so that every part of it can be plotted. When the
target depends on three or more inputs jointly, this restriction costs accuracy, and a black-box model with
post hoc explanations becomes the usual fallback. Here we present \model, a gradient boosting method whose
weak learner is a single Kolmogorov--Arnold unit $g\bigl(\sum_p \phi_p(x_p)\bigr)$ with penalized spline inner
and outer functions. Each unit is fitted in closed form by alternating penalized least squares, starting from
principal Hessian directions of the residual, and categorical and missing inputs enter the units directly.
Because a unit depends on each input only through one inner function, the fitted model has an exact functional
ANOVA decomposition into main effects, pair effects and a higher-order remainder, which one-dimensional
convolutions compute. On synthetic targets of interaction order three to six, the test error of \model{} is up to
\ordBestFold{} times lower than that of EBM, and its decomposition recovers the true variance share of every order to within
\ordShareErr. On the \ctrAllTasks{} regression tasks of the OpenML-CTR23 benchmark, it has a lower test RMSE than
EBM on \ctrAllWinsEBM{} tasks and a higher one on \ctrAllLossEBM{} (geometric-mean RMSE \ctrAllGmEBM). On these tasks the main and pair effects of the fitted model
explain a median \decRtwoMed{} of its variance, so the model reads as plots where the data allow, and the decomposition
measures the higher-order part where they do not. With Newton steps the same algorithm classifies: on \ccTasks{}
binary tasks of OpenML-CC18 its test log-loss is on par with that of EBM, and on \mcTasks{} multiclass tasks its mean
log-loss is lower than that of EBM on \mcLowerEBM{} and than that of tuned LightGBM on \mcLowerLGBM.
\end{abstract}

\section{Introduction}

Gradient-boosted decision trees remain the strongest general-purpose models for tabular data
\citep{friedman2001greedy,chen2016xgboost,ke2017lightgbm,prokhorenkova2018catboost}, and deep networks have not
displaced them on typical tabular benchmarks \citep{grinsztajn2022tree}. A boosted ensemble of thousands of trees,
however, cannot be inspected term by term. In credit, medicine and engineering, where a model's response to each
input must be checked before the model is used, inherently interpretable models are preferred to black boxes
explained after the fact \citep{rudin2019stop,caruana2015intelligible}. Post hoc attributions such as SHAP values
\citep{lundberg2017shap} describe a black box locally, but they are an approximation of the model rather than the
model itself.

Generalized additive models \citep{hastie1986gam} write the prediction as a sum of one-dimensional shape functions,
and GA$^2$M adds a selected set of two-dimensional pair surfaces \citep{lou2012intelligible,lou2013accurate}. Its
tree-boosted implementation, the explainable boosting machine (EBM) \citep{nori2019interpretml}, matches black-box
accuracy on many data sets and is the reference interpretable model for tabular regression. Neural versions keep the
same form: neural additive models \citep{agarwal2021nam}, NODE-GAM \citep{chang2022nodegam} and GAMI-Net
\citep{yang2021gaminet} learn the curves and surfaces with networks, and the functional ANOVA makes the terms
identifiable \citep{lengerich2020purifying}. All of these models stop at order two. A target that depends on three
or more inputs jointly, such as the noise of an airfoil as a function of frequency, angle and boundary-layer
thickness, is approximated by its best main-plus-pair projection, and the remainder is lost.

The Kolmogorov--Arnold representation offers a way past order two without giving up one-dimensional pieces. A
single Kolmogorov--Arnold (KA) unit $g\bigl(\sum_p \phi_p(x_p)\bigr)$ couples all of its inputs through the outer
function $g$, yet consists only of curves. Statistics has studied this form as a generalized additive model with
an unknown link \citep{breiman1985ace,horowitz2007rate,spiegel2019flexible,lin2018efficient,tutz2016generalized},
and spline networks built from several such units were fitted by backfitting \citep{coppejans2004kolmogorov} and by least squares in both layers \citep{kolodyazhniy2004fuzzy}, long before Kolmogorov--Arnold networks (KANs) brought the
form to deep learning \citep{liu2024kan}. Projection pursuit regression grows a model one nonparametric unit at a
time, but each unit sees a linear projection of the inputs \citep{friedman1981ppr,hwang1994regression}. Boosting
with KA-type learners has appeared twice: transform regression feeds each stage's output into the next stage as an
input \citep{pednault2006transform}, and GB-KAN boosts small KANs whose outer activation is fixed, so that its
interpretable configuration is additive \citep{mohr2026gbkan}.

Three things are missing for an interpretable regressor that goes beyond pairs. First, a weak learner whose
outer function is learned, so that each boosting round can add an interaction of any order. Second, a fitting
procedure for that learner in closed form, without stochastic gradient descent. Third,
a way to read the fitted model back as main effects and pair surfaces, with the higher-order part measured
rather than hidden.

\begin{figure}[t]
\centering
\includegraphics[width=\textwidth]{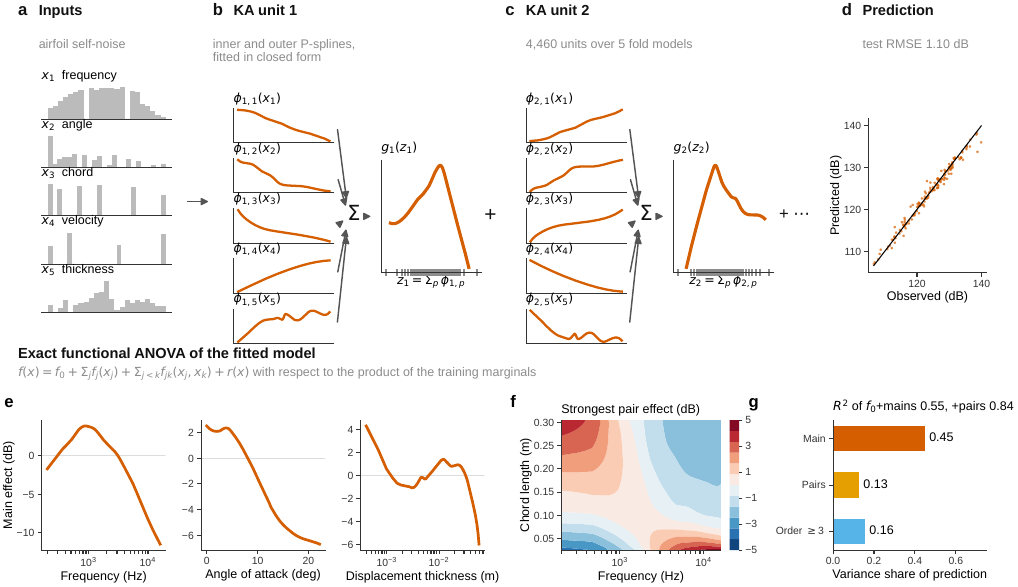}
\caption{\model{} on the airfoil self-noise task of CTR23 (fold 0). \textbf{a}, The five inputs. \textbf{b},
\textbf{c}, Two of the fitted KA units: each sums inner P-spline functions $\phi_{m,p}(x_p)$ of a few inputs and
passes the sum $z_m$ through an outer P-spline $g_m$ (ticks: deciles of $z_m$ on the training rows).
\textbf{d}, The boosted sum of all units (with the additive backbone, averaged over the five inner-fold models) on
the test rows. \textbf{e}--\textbf{g}, The exact functional ANOVA of the same model: the three largest main
effects, the strongest pair surface, and the variance share of each part on the test rows with the variance
explained by the order-$\leq$1 and order-$\leq$2 truncations. The shares do not sum to one because the parts
covary on dependent inputs.}
\label{fig:overview}
\end{figure}

Here we present \model, gradient boosting with one KA unit per round (Fig.~\ref{fig:overview}). Its inner and
outer functions are penalized B-splines (P-splines) \citep{eilers1996flexible}, and each unit is fitted in closed
form by alternating a one-dimensional fit of $g$ with a Gauss--Newton fit of all $\phi_p$ at once, started from
principal Hessian directions of the residual \citep{li1992phd}. An additive P-spline backbone with
Fellner--Schall smoothing \citep{wood2017generalized} carries the main effects. Because each unit depends on an
input only through one inner function, the fitted model has an exact functional ANOVA decomposition
\citep{hoeffding1948class,sobol2001global,hooker2007generalized} that reduces to one-dimensional convolutions. Our
contributions are:
\begin{itemize}
\item \textbf{A KA-unit boosting algorithm} with a closed-form unit fit, principal-Hessian initializations,
support pruning, an additive backbone, inner cross-validation that chooses the resolution, native
handling of categorical and missing inputs, and Newton-step forms for binary and multiclass classification
(Section~\ref{sec:method}).
\item \textbf{An exact functional ANOVA of the fitted model} into main effects, pair surfaces and a higher-order
remainder, computed per unit by one-dimensional convolutions (Section~\ref{sec:decomp}); a case study on house
prices reads one fitted model as a location map and a few curves (Section~\ref{sec:case}).
\item \textbf{Evidence on synthetic and benchmark data.} On targets of order three to six, the error of \model{} is
up to \ordBestFold{} times lower than that of EBM, and recovers the true variance share of each order to within \ordShareErr.
On the \ctrAllTasks{} tasks of OpenML-CTR23 \citep{fischer2023ctr23}, it is more accurate than EBM on
\ctrAllWinsEBM{} tasks and less accurate on \ctrAllLossEBM{}; on
\ccTasks{} binary tasks of OpenML-CC18 its log-loss is on par with EBM's, and on \mcTasks{} multiclass tasks its mean
log-loss is lower than EBM's on \mcLowerEBM{} and than tuned LightGBM's on \mcLowerLGBM{} (Section~\ref{sec:exp}).
\end{itemize}

\section{Related work}
\label{sec:related}

\paragraph{Interpretable additive models.} GAMs \citep{hastie1986gam} fit shape functions by backfitting with scatterplot smoothers, and GA$^2$M/EBM \citep{lou2012intelligible,lou2013accurate,nori2019interpretml} by cyclic boosting of shallow trees; neural variants \citep{agarwal2021nam,chang2022nodegam,yang2021gaminet} fit them by stochastic gradient
descent. \model{} contains these models as its order-$\leq$2 part (Section~\ref{sec:decomp}) and adds units of
higher order. The functional ANOVA of a fitted model \citep{hooker2007generalized} is used by
\citet{lengerich2020purifying} to make the pair terms of GA$^2$M identifiable; we use the same decomposition with
respect to the product of the training marginals to split a model of unrestricted order into its parts.

\paragraph{Unknown-link models and Kolmogorov--Arnold networks.} One KA unit is a generalized additive model with an
unknown link. Its estimation by alternating smoothing goes back to ACE \citep{breiman1985ace}; rate theory for sums
and nestings of such models is given by \citet{horowitz2007rate}; \citet{lin2018efficient} and
\citet{spiegel2019flexible} fit one unit by linearizing the link inside backfitting, with kernels and with P-splines
respectively, and \citet{tutz2016generalized} build one unit by boosting. Our unit fit follows the same
linearization, applied to every round of a boosting sequence. Spline networks with several units were fitted by backfitting \citep{coppejans2004kolmogorov} and, with piecewise-linear functions in both layers, by least squares, since such a network is linear in the parameters of each layer \citep{kolodyazhniy2004fuzzy}; KANs \citep{liu2024kan} train deep stacks of such layers by gradient-based optimization.

\paragraph{Stagewise and boosted learners.} Projection pursuit regression \citep{friedman1981ppr} and its network
form \citep{hwang1994regression} add one nonparametric unit per stage on a linear projection of the inputs. Componentwise
P-spline boosting \citep{schmid2008boosting,buhlmann2007boosting} is the special case of our learner with $g$ the
identity and one input per unit. Transform regression \citep{pednault2006transform} boosts additive stages and
chains their outputs, and GB-KAN \citep{mohr2026gbkan} boosts shallow KANs with a fixed outer activation trained by
AdamW. \model{} differs from both in its weak learner, one unit whose outer function is a learned spline, and in its
closed-form fit.

\section{\model}
\label{sec:method}

\subsection{Model}
Let $x\in\mathbb{R}^D$ and $y\in\mathbb{R}$. \model{} predicts
\begin{equation}
F(x) = f_0 + b(x) + \nu \sum_{m=1}^{M} h_m(x), \qquad h_m(x) = g_m\Bigl(\sum_{p\in S_m} \phi_{m,p}(x_p)\Bigr),
\label{eq:model}
\end{equation}
where $f_0$ is the training mean, $b(x)=\sum_p b_p(x_p)$ is an optional additive backbone, $\nu$ is the learning
rate, and $S_m$ is the support of unit $m$. Each inner function $\phi_{m,p}$ is a cubic B-spline on quantile knots
of $x_p$, centered on the training rows, and each outer function $g_m$ is a cubic B-spline on quantile knots of the
unit's inner sum $z_m = \sum_p \phi_{m,p}(x_p)$. Inputs with few distinct values get a knot at every value.
With $g_m$ the identity, a unit is an additive model; with a nonlinear $g_m$ it couples all inputs in $S_m$, so a
sum of units can represent interactions of any order. A product $x_1x_2$ over a domain where both signs occur,
for instance, needs two units, $\tfrac14[(x_1+x_2)^2-(x_1-x_2)^2]$.

\subsection{Fitting one unit in closed form}
\label{sec:unit}
Round $m$ fits $h_m$ to the residual $r = y - F_{m-1}(x)$. Write the inner design of input $p$ as the centered
B-spline matrix $\mathbf{B}_p$, with coefficients $\bm\alpha_p$, so that $z = \sum_p \mathbf{B}_p\bm\alpha_p$, and
the outer design as $\mathbf{G}(z)$ with coefficients $\bm\gamma$. The unit minimizes the penalized squared error
\begin{equation}
\frac1N\bigl\|r - \mathbf{G}(z)\bm\gamma\bigr\|^2 + \lambda_{\rm out}\,\bm\gamma^\top\mathbf{P}\bm\gamma
 + \lambda_{\rm in}\sum_p \bm\alpha_p^\top\mathbf{P}_p\bm\alpha_p,
\end{equation}
with second-difference penalties $\mathbf{P}$, $\mathbf{P}_p$ \citep{eilers1996flexible}. We alternate two
closed-form steps, rescaling $z$ to unit variance after each:
\begin{itemize}
\item \emph{$g$-step.} With $z$ fixed, $\bm\gamma$ solves a one-dimensional P-spline regression of $r$ on $z$, with
knots at the quantiles of $z$.
\item \emph{$\phi$-step.} With $g$ fixed, linearize $g(z+\delta)\approx g(z)+g'(z)\,\delta$. The new inner
coefficients solve one weighted additive P-spline regression,
\begin{equation}
\min_{\bm\alpha}\ \frac1N\sum_{i} g'(z_i)^2\Bigl(\tilde r_i - \sum_p [\mathbf{B}_p\bm\alpha_p]_i\Bigr)^2 +
\lambda_{\rm in}\sum_p\bm\alpha_p^\top\mathbf{P}_p\bm\alpha_p,\qquad \tilde r_i = z_i + \frac{r_i-g(z_i)}{g'(z_i)},
\end{equation}
a Gauss--Newton step for all inner functions at once.
\end{itemize}
Each step is a small symmetric positive definite solve. Because B-splines have local support, every row has only
four nonzero basis values per input, so the normal equations of the $\phi$-step cost $O(N(4|S|)^2)$ rather than
$O(NK^2)$ for $K$ inner coefficients. We keep the iterate with the lowest training loss.

\paragraph{Initialization.} The alternation needs a starting $z$. We try the additive fit of $r$ (the unit with
$g$ the identity) and the $n_h=3$ leading principal Hessian directions of the residual \citep{li1992phd}, the
eigenvectors $v$ of $\E[r\,(x-\bar x)(x-\bar x)^\top]$ with the largest $|$eigenvalue$|$. A direction $v$ starts
the unit at $\phi_p(x_p)=v_p x_p$: along it a quadratic $g$ exposes pairwise structure, as in the product above.
All initializations run on a random window of $\max(1000, 0.15N)$ training rows; the best one is then refitted on
all rows.

\paragraph{Support.} Each unit draws $\mathrm{round}(c\cdot\min(D,m_{\max}))$ inputs at random, where the column
fraction $c$ is chosen by cross-validation and $m_{\max}=12$ caps the support of units on wide data. After the
first fit, support pruning drops the inputs whose inner function carries less than 15\% of the unit's inner variance
and refits the unit on the rest. Cross-validation decides whether to prune: pruning keeps units small and smooth,
while a conjunction of several binary inputs needs inputs that carry little variance on their own. On the CTR23
tasks of Section~\ref{sec:reading} a unit uses \supMin{} to \supMax{} inputs on average.

\subsection{Additive backbone}
Before the units, an optional backbone fits an additive P-spline model to $y-f_0$ with one smoothing parameter per
input, chosen by the generalized Fellner--Schall update \citep{wood2017generalized}. Its resolution (16 to 256 knots)
is chosen on held-out rows. The backbone takes the main effects in one solve, so that units start on the
interaction structure; whether it helps is decided by cross-validation (below).

\subsection{Model selection}
\label{sec:select}
Five-fold cross-validation inside the training data chooses the backbone (on or off), the unit resolution (16 or 32
interior knots for $\phi$ and $g$), the column fraction ($c\in\{0.5, 1\}$) and support pruning (on or off). The
candidates are raced: all run on the first fold, where a candidate is stopped at 100 and 200 units if its validation
squared errors exceed the leader's by more than two standard errors of the paired difference, and three finalists
continue on the other folds. A race decides on one fold and early in the fit, and the effect of pruning can show only
later, so two races run on the same folds: one over the eight candidates with pruning and one over all sixteen. The
winner of the second race replaces that of the first when its cross-validated error is lower. Each
fold model is early-stopped on its own fold with a patience of 50 units, and the prediction is the average of the
five fold models of the chosen candidate. The learning rate is $\nu=0.3$ and $\lambda_{\rm in}=0.1$,
$\lambda_{\rm out}=10^{-3}$ throughout (Appendix~\ref{app:settings}).

\subsection{Categorical and missing inputs}
\label{sec:cat}
A categorical input is one input of the units, whatever its number of levels: its inner function is a table with one
value per training level (one-hot basis, with missing as a level of its own). Inside units the level values are shrunk toward zero, as in empirical Bayes encodings of high-cardinality categorical inputs \citep{micci2001preprocessing}: the $\phi$-step adds to the diagonal of each level the
weight of $\kappa=10$ average rows, so that a level with $n$ training rows keeps about $n/(n+\kappa)$ of its own
signal, independently of the scale of the target. In the backbone the level values carry a ridge penalty whose weight
the Fellner--Schall update estimates, as for a random effect. For the principal Hessian directions a categorical
input is represented by the shrunk mean residual of its level, recomputed for every unit. A numeric input with
missing values gets one more coefficient in its inner function, the value $\phi_p(\textsc{nan})$, next to the spline
of its observed values; the principal Hessian directions see the training median. A level or a missing value not seen
in training contributes the reference value of the inner function. Because a categorical input occupies one slot of
a unit, a unit can combine, say, a game, a graphics card and a resolution, which a one-hot encoding with a cap of
$m_{\max}$ inputs per unit cannot.

\subsection{Classification}
\label{sec:clf}
For a binary target $y\in\{0,1\}$, $F$ in~\eqref{eq:model} is the log-odds and each round takes a Newton step on the
log-loss, as in LogitBoost \citep{friedman2000additive}. With $p=\sigma(F)$, the unit is fitted by weighted least
squares to the working response $(y-p)/h$, clipped to $|\cdot|\le 4$, with row weights $h=p(1-p)$. Both closed-form
steps of Section~\ref{sec:unit} carry these weights, the additive initialization becomes a weighted additive fit, and
the principal Hessian directions use the gradient $y-p$. The base score is the log-odds of the training frequency;
the backbone is a logistic additive P-spline model fitted by penalized iteratively reweighted least squares with
Fellner--Schall smoothing in every iteration; and early stopping and the inner cross-validation use the validation
log-loss. The decomposition of Section~\ref{sec:decomp} applies unchanged to the log-odds. With the squared loss
every computation reduces to the regression algorithm.

With $K\ge3$ classes, $F$ is a vector of $K$ logits and the class probabilities are its softmax. Each round adds one
unit per class: with $p$ the softmax of $F$ at the start of the round, the unit of class $k$ is fitted to the working
response $(y_k-p_k)/h_k$, clipped as above, with row weights $h_k=\tfrac{K}{K-1}\,p_k(1-p_k)$, the diagonal Newton
step of the $K$-class gradient boosting of \citet{friedman2001greedy}. The base score is the centred log of the class
frequencies, the backbone has one additive predictor per class fitted by the same Newton steps, and early stopping and
the inner cross-validation use the multinomial log-loss. Each logit is a sum of backbone terms and units, so its
decomposition is exact as well.

\begin{algorithm}[t]
\caption{\model{} (one fold model; the inner cross-validation of Section~\ref{sec:select} runs it per candidate)}
\label{alg:kau}
\begin{algorithmic}[1]
\State $F\gets f_0$; \textbf{if} backbone: fit $b$ by Fellner--Schall and set $F\gets F+b$
\For{$m=1,\dots,M_{\max}$}
  \State $r\gets y-F$; draw the candidate support $S$ of $\mathrm{round}(c\min(D,m_{\max}))$ inputs
  \State initializations: additive fit of $r$; $n_h$ principal Hessian directions of $r$
  \State run the $g$-step/$\phi$-step alternation from each on a row window; prune inputs below 15\% of the inner variance (if chosen)
  \State refit the best initialization on all rows $\Rightarrow h_m$; \ $F\gets F+\nu h_m$
  \State stop when the validation error has not improved for 50 units; keep the best $M$
\EndFor
\end{algorithmic}
\end{algorithm}

\subsection{Computation}
The $\phi$-step normal equations are accumulated from the four nonzero basis values per input and row, with the
centering added back as a rank-two correction; the whole alternation of a unit's initializations runs in one
compiled call \citep{lam2015numba}. The additive initialization reuses one Cholesky factor across rounds, since its
matrix does not change. All runs in this paper use one CPU thread per fit.

\section{Exact functional ANOVA of a fitted model}
\label{sec:decomp}

Let $\mu=\mu_1\otimes\cdots\otimes\mu_D$ be the product of the training marginals. The functional ANOVA
\citep{hoeffding1948class,sobol2001global,hooker2007generalized} writes any $F$ as
\begin{equation}
F(x) = f_0 + \sum_j f_j(x_j) + \sum_{j<k} f_{jk}(x_j,x_k) + \rho(x),
\end{equation}
with $f_0=\E_\mu F$, $f_j(x_j)=\E_\mu[F\mid x_j]-f_0$, $f_{jk}=\E_\mu[F\mid x_j,x_k]-f_j-f_k-f_0$, and $\rho$ the
interactions of order three and higher. For a black box these conditional expectations need integration over $D-1$
or $D-2$ inputs. For \model{} they are one-dimensional. A unit depends on $x_L$ ($L=\{j\}$ or $\{j,k\}$) only through
$t=\sum_{p\in L}\phi_p(x_p)$, so
\begin{equation}
\E_\mu\bigl[h(X)\mid x_L\bigr] = H_L(t), \qquad H_L(t)=\E\, g\Bigl(t+\sum_{p\in S\setminus L}\phi_p(X_p)\Bigr),
\label{eq:hl}
\end{equation}
and the distribution of $\sum_{p\in S\setminus L}\phi_p(X_p)$ is the convolution of the distributions of the
$\phi_p(X_p)$. Each marginal is the training column (its distinct values with their frequencies, or 2{,}000
equally weighted quantiles when there are more); we bin each $\phi_p(X_p)$ linearly on a common grid of 2{,}048 steps
over the range of the unit's inner sum, convolve, and correlate with $g$ on the grid. This is exact up to the
$O(\text{step}^2)$ binning error, with no sampling. When nothing is left to average ($L=S$), $H_L=g$, so units with
at most two inputs contribute nothing to $\rho$. The backbone contributes main effects only, and the fold average
of the prediction decomposes as the average of the fold decompositions. The parts sum to the prediction at every
point: on the CTR23 test rows the largest deviation is $\decCheck$ in the units of the target.

We summarize a decomposition on test rows by the variance share of each part, such as
$\Var(\sum_j f_j)/\Var(F)$ for the main effects, and by the $R^2$ with respect to $F$ of the order-1 truncation
$f_0+\sum_j f_j$ and the order-2 truncation $f_0+\sum_j f_j+\sum_{j<k}f_{jk}$. The parts are uncorrelated under $\mu$; on dependent test
rows they can covary, so the shares need not sum to one.

\section{Experiments}
\label{sec:exp}

\subsection{Setup}
\paragraph{Development and test data.} We designed \model{} on seven regression data sets (California housing, CCPP,
abalone, airfoil self-noise, concrete, diabetes, energy efficiency; outer five-fold cross-validation). The benchmark
is the OpenML-CTR23 suite \citep{fischer2023ctr23,vanschoren2014openml}: 35 regression tasks
with 517 to 72{,}000 rows and 5 to 116 inputs, evaluated on the ten folds of the first repeat of the official splits.
Five CTR23 tasks (abalone, airfoil self-noise, California housing, concrete, energy efficiency) also belong to the
development set; we report results on all 35 tasks and on the 30 others. For CTR23 the support cap $m_{\max}=12$ is added, which
leaves every fit with at most 12 inputs unchanged. Table~\ref{tab:dev} gives the development results.

\paragraph{Baselines.} EBM \citep{nori2019interpretml} with its default settings (interactions on), the reference
interpretable model; and LightGBM \citep{ke2017lightgbm} tuned over a 12-point grid by inner five-fold
cross-validation with early stopping, as a black-box reference. All models see the same folds; EBM and LightGBM
use their own handling of categorical and missing values, and \model{} uses that of Section~\ref{sec:cat}. Numeric
inputs are standardized on the training part. Every fit runs on one CPU thread; we compare test RMSE with paired
$t$-tests over the folds ($p<0.05$).

\subsection{Interactions beyond pairs}
We first test whether \model{} learns interactions of order higher than two, and whether its decomposition
recovers them. Targets on $D=8$ inputs, $x\sim U(-1.5,1.5)^8$, depend on the first $k$ inputs through an exponential
or a sine of an additive index, $f=\exp(s)$ or $\sin(2s)$ with $s=k^{-1/2}\sum_{p\le k}(x_p+\tfrac12\sin 2x_p)$, or
through the product $f=\prod_{p\le k}x_p$; 3{,}000 training and 2{,}000 test rows, noise of $0.1\,\mathrm{sd}(f)$,
three seeds. Each target is a product of independent one-input factors, so its true Hoeffding shares follow in closed
form (Appendix~\ref{app:shares}); Table~\ref{tab:order} lists all values.

\begin{figure}[t]
\centering
\includegraphics[width=\textwidth]{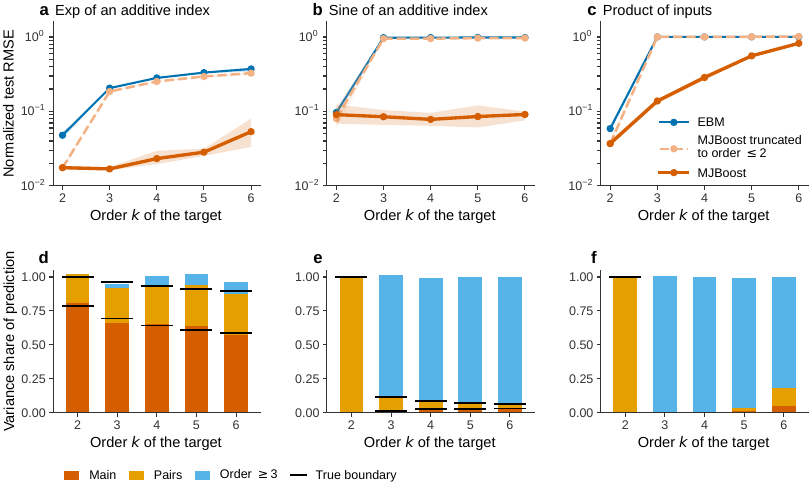}
\caption{Interaction-order sweep. \textbf{a}--\textbf{c}, Normalized test RMSE (RMSE$/\mathrm{sd}(f)$ against the
noise-free target) of \model, of \model{} truncated to its order-$\leq$2 functional ANOVA part, and of EBM, as a
function of the order $k$ of the target; lines are means over three seeds, shading their range.
\textbf{d}--\textbf{f}, Variance shares of the main, pair and order-$\geq$3 parts of the fitted model on the test
rows (bars, mean over seeds) and the true Hoeffding shares (black lines: boundaries between the orders).}
\label{fig:order}
\end{figure}

For $k\ge3$ EBM fails on every family: its error stays close to that of the best order-$\le$2 approximation
(Fig.~\ref{fig:order}a--c). \model{} truncated to its own order-$\le$2 part fails in the same way, while the full
model does not: at $k=3$ on the exponential family the normalized test RMSE is \ordExpThreeKAU{} for \model,
\ordExpThreeUpTwo{} for its truncation and \ordExpThreeEBM{} for EBM. The accuracy therefore comes from the units of
order three and higher. Up to $k=6$ for the index families and $k=4$ for the product, the variance shares of the
fitted model's parts match the true shares to within \ordShareErr{} (Fig.~\ref{fig:order}d--f), and main and pair
terms attributed to the irrelevant inputs carry at most \ordSpurMax{} of the variance. On a pure product of five or six
inputs, whose whole variance is of the top order, \model{} still reaches \ordProdFive{} and \ordProdSix, while EBM
stays at \ordProdEBM, the error of the constant mean.

\subsection{Accuracy on OpenML-CTR23}
\begin{figure}[!tbp]
\centering
\includegraphics[width=\textwidth]{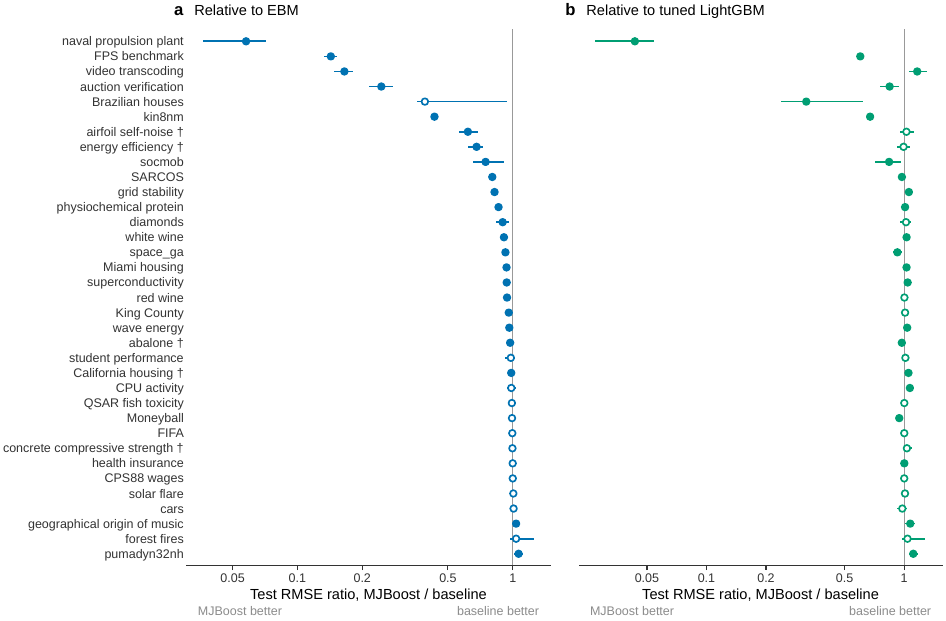}
\caption{Per-task CTR23 test RMSE of \model{} relative to EBM (\textbf{a}) and to tuned LightGBM (\textbf{b}). Points:
ratio of the ten-fold mean RMSEs; whiskers: 95\% interval of the geometric-mean fold ratio; filled points: paired
$t$-test $p<0.05$. Tasks sorted by the ratio to EBM; $\dagger$: tasks that overlap the development data. Per-task
values are in Table~\ref{tab:pertask}.}
\label{fig:ctr23}
\end{figure}

Table~\ref{tab:ctr23} and Fig.~\ref{fig:ctr23} compare test RMSE on the 35 tasks. \model{} is more accurate than
EBM on \ctrAllWinsEBM{} tasks, statistically tied on \ctrAllTiesEBM{} and less accurate on \ctrAllLossEBM{}; its
geometric-mean RMSE is \ctrAllGmEBMabs{} lower. On the \ctrHeldTasks{} tasks outside the development data the
counts are \ctrHeldWinsEBM/\ctrHeldTiesEBM/\ctrHeldLossEBM{} with a \ctrHeldGmEBMabs{} lower geometric-mean RMSE.
Across the three models the mean ranks are \ctrAllRankKAU{} for \model, \ctrAllRankEBM{} for EBM and
\ctrAllRankLGBM{} for tuned LightGBM (Friedman $p=\ctrAllFriedman$). The largest gains over EBM occur where the
targets depend on several inputs jointly, such as the naval propulsion plant, auction verification, Brazilian
houses, video transcoding and kin8nm tasks, and on tasks with high-cardinality categorical inputs, such as the FPS
benchmark, where a categorical input enters a unit as one input (Section~\ref{sec:cat}). Against tuned LightGBM, a
black box, \model{} is significantly more accurate on \ctrAllWinsLGBM{} tasks (\ctrBeatLGBMlist), tied on
\ctrAllTiesLGBM{} and less accurate on \ctrAllLossLGBM; its geometric-mean RMSE is \ctrAllGmLGBMabs{}
\ctrAllGmLGBMdir, an average to which the tasks with the largest gains contribute most.

\begin{table}[t]
\centering
\caption{CTR23 test RMSE of \model{} against EBM and tuned LightGBM. W/T/L: tasks with significantly lower,
statistically equal and significantly higher RMSE for \model{} (paired $t$-test over the 10 folds, $p<0.05$). RMSE
ratio: geometric mean over tasks of the ratio of fold-mean RMSEs, minus one. Mean rank among the three models
(1 = lowest RMSE).}
\label{tab:ctr23}
\small
\resizebox{\textwidth}{!}{\begin{tabular}{lccccc}
\toprule
 & \multicolumn{2}{c}{vs.\ EBM} & \multicolumn{2}{c}{vs.\ tuned LightGBM} & Mean rank \\
\cmidrule(lr){2-3}\cmidrule(lr){4-5}
Tasks & W/T/L & RMSE ratio & W/T/L & RMSE ratio & MJBoost / EBM / LightGBM \\
\midrule
All 35 tasks & 21/12/2 & $-$29.1\% & 10/13/12 & $-$13.2\% & 1.80 / 2.71 / 1.49 \\
30 tasks outside the development data & 17/11/2 & $-$30.9\% & 9/10/11 & $-$15.4\% & 1.83 / 2.67 / 1.50 \\
\bottomrule
\end{tabular}}
\end{table}

\subsection{Reading the fitted models}
\label{sec:reading}
\begin{figure}[!tbp]
\centering
\includegraphics[width=\textwidth]{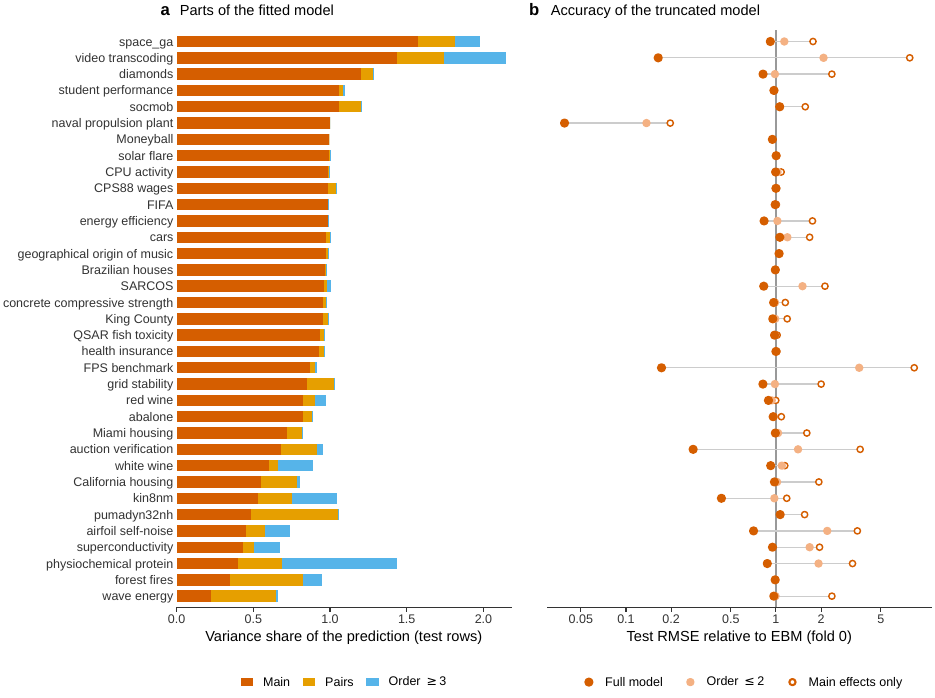}
\caption{Functional ANOVA of \model{} on \decTasks{} CTR23 tasks (fold 0). \textbf{a}, Variance shares of the main,
pair and order-$\geq$3 parts of the prediction on the test rows; the shares can exceed one in sum when the parts
covary on dependent inputs. \textbf{b}, Test RMSE of the full model and of its order-$\leq$2 and order-1 truncations,
relative to EBM on the same fold.}
\label{fig:decomp}
\end{figure}

How much of a fitted \model{} lies beyond pairs? We decomposed the fold-0 models of all \decTasks{} CTR23 tasks
(Fig.~\ref{fig:decomp}). The main effects carry a median
\decMainMed{} of the prediction's variance, the pairs \decPairMed{} and the order-$\geq$3 part \decRestMed;
main and pair effects together reproduce a median \decRtwoMed{} of the variance. On most tasks, then, a fitted
\model{} can be read as a set of curves and surfaces, like an EBM. The higher-order part is small but not idle:
dropping it raises the test RMSE by a median \decUpTwoLossMed, and by much more on four tasks: \decUpTwoTop. The order-$\le$2 part of \model{} alone is a GA$^2$M as accurate as EBM: it has a lower test
RMSE than EBM on \decBeatEBMtwo{} of the \decTasks{} tasks, and the full model on \decBeatEBMfull.

Fig.~\ref{fig:overview}e--g shows the decomposition of the airfoil model. The frequency effect peaks near 1~kHz and
the noise falls with angle of attack; the strongest pair couples frequency with chord length, raising the noise of
long chords at low frequencies and lowering it at high frequencies, consistent with a spectral peak that moves to
lower frequencies as the longer chord grows a thicker boundary layer. On this task the main and pair effects explain \figOneRtwo{} of the
variance, yet the higher-order part decides the accuracy: the test RMSE is \figOneRMSE~dB for the full model,
\figOneRMSEtwo~dB for its order-$\leq$2 truncation and \figOneRMSEebm~dB for EBM.

\subsection{Case study: house prices in King County}
\label{sec:case}
\begin{figure}[!tbp]
\centering
\includegraphics[width=\textwidth]{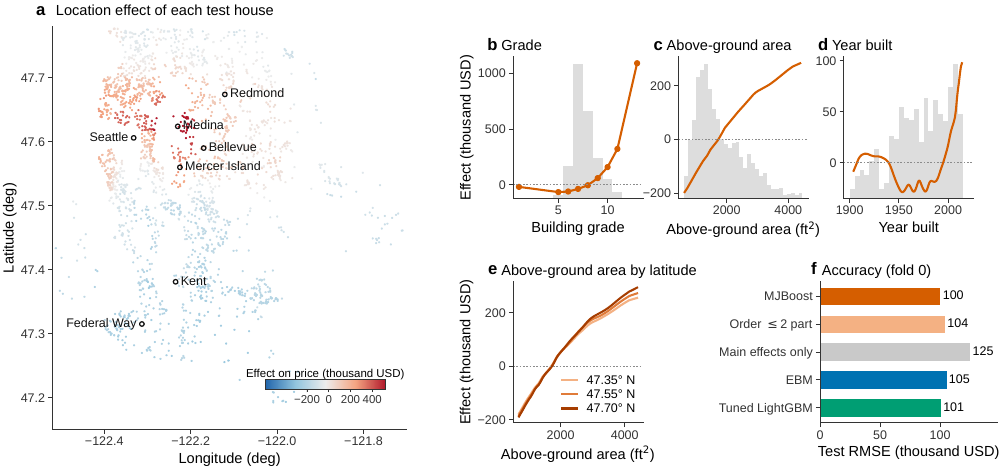}
\caption{Case study: the fold-0 \model{} model of King County house prices (CTR23), read through its functional
ANOVA. \textbf{a}, Location effect of each test house, the sum of all main and pair terms of latitude, longitude and
zip code (thousand USD; gray: the median house; color scale clipped at the 99th percentile of the deviation from it);
circles mark towns. \textbf{b}--\textbf{d}, Main effects of building grade, above-ground area and year built (lines) over
the distribution of the test houses (gray bars). \textbf{e}, Effect of the above-ground living area at three
latitudes: its main effect plus the area $\times$ latitude pair term. \textbf{f}, Test RMSE on the same fold of the
full model, of its order-$\leq$2 and order-1 truncations, of EBM and of tuned LightGBM.}
\label{fig:case}
\end{figure}

Fig.~\ref{fig:case} reads one fitted model in full: the fold-0 model of the King County task of CTR23, sale prices of
\caseN{} houses around Seattle sold in 2014 and 2015, with \caseD{} inputs among which the 70 zip codes enter as one
categorical input (Section~\ref{sec:cat}). Main and pair effects reproduce \caseRtwo{} of the variance of its
prediction, so the model reads almost entirely as curves and surfaces, each an exact part of the fitted model (parts
and remainder sum to the prediction to within $\caseCheck$~USD). Summing all main and pair terms of latitude,
longitude and zip code gives every house a location effect, from \caseLocLo{} to \caseLocHi{} thousand USD between
the 1st and 99th percentiles (Fig.~\ref{fig:case}a). It is highest in Seattle and across Lake Washington in Medina,
Mercer Island and Bellevue, and lowest in the south of the county around Kent and Federal Way. Going from
building grade 7, the most common, to grade 11 adds \caseGradeSevenEleven{} thousand USD on a convex curve
(Fig.~\ref{fig:case}b); the effect of the above-ground area, the input through which size mainly enters, rises by
about \caseAboveSlope{} USD per square foot (Fig.~\ref{fig:case}c); and the year-built effect averages $-$\caseYrMid{} thousand USD for houses built between 1950
and 1990 and +\caseYrNew{} thousand USD for those built since 2000 (Fig.~\ref{fig:case}d). The strongest interaction
of size is with latitude: enlarging the above-ground area from 1{,}500 to 4{,}000~ft$^2$ adds \caseAboveNorth{}
thousand USD at 47.70$^\circ$N and \caseAboveSouth{} thousand USD at 47.35$^\circ$N (Fig.~\ref{fig:case}e).
The full model has a test RMSE of \caseRMSE{} thousand USD on this fold, below tuned LightGBM (\caseRMSElgbm) and
EBM (\caseRMSEebm). Truncated to its readable parts, it keeps most of that accuracy: \caseRMSEtwo{} thousand USD,
still below EBM, whereas the main effects alone give \caseRMSEmain{} (Fig.~\ref{fig:case}f).

\subsection{Ablation}
Table~\ref{tab:ablation} removes one component at a time on the development data. Without units (the additive
backbone alone, a P-spline GAM) the RMSE rises by \ablBB{} on average; with units restricted to two inputs, the
order of EBM's GA$^2$M, by \ablMtwo; without the principal-Hessian initializations by \ablHess; and with support
pruning always on (the race over the eight candidates with pruning alone) by \ablPrune. The units of
higher order matter most on airfoil self-noise and energy efficiency, the two development sets with the strongest
interactions.

\begin{table}[t]
\centering
\caption{Ablation on the development data (outer five-fold test RMSE). First row: \model{}; other rows: change of
the fold-mean RMSE when one component is removed; $^{*}$: paired $t$-test $p<0.05$ over the five folds.}
\label{tab:ablation}
\small
\setlength{\tabcolsep}{4pt}
\resizebox{\textwidth}{!}{\begin{tabular}{lrrrrrrrr}
\toprule
Variant & California & ccpp & abalone & airfoil & concrete & diabetes & energy & Mean \\
\midrule
MJBoost (RMSE) & 0.4462 & 3.281 & 2.075 & 1.311 & 4.121 & 54.68 & 0.3307 & -- \\
\midrule
Additive backbone only (no units) & +23.1\%$^{*}$ & +12.0\%$^{*}$ & +2.7\%$^{*}$ & +248.0\%$^{*}$ & +29.8\%$^{*}$ & $-$0.2\% & +214.5\%$^{*}$ & +75.7\% \\
Units with at most two inputs & +5.1\%$^{*}$ & +2.9\%$^{*}$ & +2.0\% & +64.7\%$^{*}$ & +2.8\% & $-$0.1\% & +31.6\%$^{*}$ & +15.6\% \\
No principal-Hessian initializations & $-$1.9\%$^{*}$ & +0.1\% & +2.0\% & +15.5\%$^{*}$ & +3.7\% & +0.3\% & +9.2\%$^{*}$ & +4.1\% \\
Support pruning always on & +0.0\% & +0.7\% & +0.0\% & +3.2\% & +0.0\% & +0.0\% & +9.3\%$^{*}$ & +1.9\% \\
\bottomrule
\end{tabular}}
\end{table}

\subsection{Computational cost}
Over the whole CTR23 run (35 tasks $\times$ 10 folds, one thread per fit) \model{} used \cpuTotalKAU{} CPU-hours,
EBM \cpuTotalEBM{} and tuned LightGBM \cpuTotalLGBM: \model{} costs \cpuRatioEBM{} times EBM and \cpuRatioLGBM{} times
tuned LightGBM, including both races of its inner cross-validation. The decomposition of a fitted model takes seconds to minutes on
the tasks of Fig.~\ref{fig:decomp}.

\subsection{Classification}
\label{sec:cc18}
The classification form of \model{} (Section~\ref{sec:clf}) runs with the CTR23 setting on the \ccTasks{} binary
tasks of OpenML-CC18 \citep{bischl2021openml} with at most 200 inputs (the three others have 500 to 1{,}776 inputs):
\ccRowsMin{} to \ccRowsMax{} rows, \ccDMin{} to \ccDMax{} inputs, ten folds of the official splits. The baselines are the EBM classifier with its
defaults and LightGBM tuned as for CTR23 on the inner log-loss. On the test log-loss \model{} is on par with EBM
(Table~\ref{tab:cc18}): its mean log-loss is lower on \ccLowerEBM{} of the \ccTasks{} tasks, the median ratio is
\ccMedRatioEBM, and the mean ranks are \ccRankKAU{} for \model, \ccRankEBM{} for EBM and \ccRankLGBM{} for tuned
LightGBM (Friedman $p=\ccFriedman$). The mean AUC is \ccAucKAU, \ccAucEBM{} and \ccAucLGBM. \model{} used
\ccCpuKAU{} CPU-hours, EBM \ccCpuEBM{} and tuned LightGBM \ccCpuLGBM. Table~\ref{tab:cc18pertask} lists every task.

For more than two classes we use the \mcTasks{} multiclass tasks of CC18 with at most 30 inputs and 5{,}500 rows
(\mcRowsMin{} to \mcRowsMax{} rows, \mcDMin{} to \mcDMax{} inputs, \mcKMin{} to \mcKMax{} classes) and the softmax form of
Section~\ref{sec:clf}, against EBM and LightGBM tuned on the multiclass log-loss, again over the ten official folds.
\model{} has the lower mean test log-loss on \mcLowerEBM{} of the \mcTasks{} tasks against EBM (median ratio
\mcMedRatioEBM; W/T/L \mcWinsEBM/\mcTiesEBM/\mcLossEBM) and on \mcLowerLGBM{} against tuned LightGBM (median ratio
\mcMedRatioLGBM; \mcWinsLGBM/\mcTiesLGBM/\mcLossLGBM); the mean ranks are \mcRankKAU{} for \model, \mcRankEBM{} for EBM and
\mcRankLGBM{} for tuned LightGBM (Friedman $p=\mcFriedman$), and the mean accuracy is \mcAccKAU, \mcAccEBM{} and
\mcAccLGBM. The largest gains are on balance-scale (\mcGainBalance{} against EBM), whose class compares the products of weight
and distance on the two sides, and on car (\mcGainCar), whose classes come from a hierarchical decision model over
its six inputs. \model{} used \mcCpuKAU{}
CPU-hours, EBM \mcCpuEBM{} and tuned LightGBM \mcCpuLGBM. Table~\ref{tab:mcpertask} lists every task.

\begin{table}[t]
\centering
\caption{OpenML-CC18 binary and multiclass tasks: test log-loss and AUC (multiclass: macro one-vs-rest AUC) of
\model{} against EBM and tuned LightGBM. W/T/L: tasks with significantly better, statistically equal and significantly
worse values for \model{} (paired $t$-test over the 10 folds, $p<0.05$). Lower mean: tasks on which the ten-fold
mean log-loss of \model{} is lower. Median ratio: median over tasks of the ratio of the ten-fold mean log-losses,
minus one.}
\label{tab:cc18}
\small
\begin{tabular}{llcccc}
\toprule
 & & \multicolumn{3}{c}{Log-loss} & AUC \\
\cmidrule(lr){3-5}\cmidrule(lr){6-6}
Tasks & \model{} vs. & W/T/L & Lower mean & Median ratio & W/T/L \\
\midrule
Binary & EBM & 3/21/8 & 17 of 32 & $-$0.1\% & 4/23/5 \\
 & tuned LightGBM & 2/19/11 & 11 of 32 & +2.5\% & 3/22/7 \\
\midrule
Multiclass & EBM & 4/4/3 & 8 of 11 & $-$5.1\% & 5/5/1 \\
 & tuned LightGBM & 7/3/1 & 7 of 11 & $-$7.8\% & 5/3/3 \\
\bottomrule
\end{tabular}
\end{table}

\section{Conclusion}
\model{} boosts one Kolmogorov--Arnold unit per round, fits each unit in closed form, and decomposes the fitted model
exactly into main effects, pair surfaces and a measured higher-order part. It learns interactions of order three and
higher that additive and GA$^2$M models cannot represent, is more accurate than EBM on CTR23, and
remains readable as curves and surfaces wherever the data allow. Weighted by Newton steps, the same unit fit gives a
binary classifier on par with EBM on OpenML-CC18 and a multiclass classifier whose log-loss is lower than that of EBM
and of tuned LightGBM on most of its multiclass tasks.

\section*{Code availability}
\model{} is available as the Python package \texttt{mjboost} on PyPI (\url{https://pypi.org/project/mjboost/};
\texttt{pip install mjboost}), with a scikit-learn interface and the functional ANOVA of Section~\ref{sec:decomp};
version 0.2.0 produced the multiclass results of Section~\ref{sec:cc18}. The scripts that produce every number, table
and figure of this paper will be released with the paper.

\appendix
\renewcommand{\thetable}{A\arabic{table}}
\setcounter{table}{0}
\section{Settings}
\label{app:settings}
All runs use one setting:
learning rate $\nu=0.3$; at most 4{,}000 units; early-stopping patience 50; $\lambda_{\rm in}=0.1$,
$\lambda_{\rm out}=10^{-3}$ (per-row scale); cubic B-splines, a knot at every distinct value for inputs with few
values; inner cross-validation with five folds over the backbone (Fellner--Schall additive P-splines, resolution
16 to 256 knots with at most 1{,}100 basis functions, or none), the unit resolution (16 or 32 interior knots for both
$\phi$ and $g$), the column fraction (0.5 or 1) and support pruning (on or off); two races on the same folds, over
the eight candidates with pruning and over all sixteen, each on the first fold with stops at 100 and 200 units (two
standard errors) and three finalists, the winner of the second kept when its cross-validated error is lower;
prediction by the average of the five fold models;
$n_h=3$ principal-Hessian initializations; three alternation steps per unit; support pruning, when chosen, at 15\% of the inner
variance; first fits on a window of $\max(1000, 0.15N)$ rows and a refit of the best initialization on all rows;
support cap $m_{\max}=12$; categorical shrinkage $\kappa=10$ rows.
The CC18 runs use the same setting with the logistic loss of Section~\ref{sec:clf} (binary tasks) or its softmax
form (multiclass tasks): working response clipped at $|\cdot|\le4$ and at most four iteratively reweighted
least-squares steps for the backbone.

\section{Development benchmark}
\begin{table}[h]
\centering
\caption{Development data: outer five-fold test RMSE of \model{} (the final setting) and EBM; $p$: paired $t$-test.}
\label{tab:dev}
\small
\begin{tabular}{lrrrr}
\toprule
Data set & MJBoost & EBM & vs.\ EBM & $p$ \\
\midrule
California housing & 0.4462 & 0.4625 & $-$3.5\% & 0.019 \\
ccpp & 3.281 & 3.342 & $-$1.8\% & 0.026 \\
abalone & 2.075 & 2.135 & $-$2.8\% & 0.012 \\
airfoil self-noise & 1.311 & 1.931 & $-$32.1\% & 0.000 \\
concrete & 4.121 & 4.116 & +0.1\% & 0.940 \\
diabetes & 54.68 & 54.57 & +0.2\% & 0.779 \\
energy efficiency & 0.3307 & 0.4341 & $-$23.8\% & 0.001 \\
\bottomrule
\end{tabular}
\end{table}

\section{Interaction-order sweep}
\begin{table}[h]
\centering
\caption{Interaction-order sweep (means over three seeds): normalized test RMSE and variance shares
(main/pair/order $\geq$3) of the fitted model and of the target.}
\label{tab:order}
\small
\begin{tabular}{llrrrll}
\toprule
Family & $k$ & MJBoost & MJBoost$_{\le 2}$ & EBM & Shares (main/pair/$\ge$3) & True shares \\
\midrule
exp-sum & 2 & 0.018 & 0.017 & 0.048 & 0.81/0.22/0.00 & 0.79/0.21/0.00 \\
 & 3 & 0.017 & 0.185 & 0.206 & 0.66/0.26/0.03 & 0.69/0.27/0.03 \\
 & 4 & 0.023 & 0.253 & 0.282 & 0.65/0.29/0.06 & 0.64/0.29/0.06 \\
 & 5 & 0.028 & 0.295 & 0.332 & 0.64/0.31/0.08 & 0.61/0.30/0.09 \\
 & 6 & 0.053 & 0.327 & 0.372 & 0.57/0.30/0.09 & 0.59/0.31/0.10 \\
\midrule
sin-sum & 2 & 0.091 & 0.080 & 0.097 & 0.00/1.00/0.00 & 0.00/1.00/0.00 \\
 & 3 & 0.085 & 0.948 & 0.977 & 0.02/0.10/0.89 & 0.02/0.10/0.88 \\
 & 4 & 0.078 & 0.951 & 0.980 & 0.03/0.07/0.90 & 0.03/0.06/0.91 \\
 & 5 & 0.085 & 0.965 & 0.985 & 0.03/0.05/0.92 & 0.03/0.04/0.93 \\
 & 6 & 0.091 & 0.970 & 0.984 & 0.03/0.04/0.93 & 0.03/0.03/0.94 \\
\midrule
product & 2 & 0.037 & 0.037 & 0.059 & 0.00/1.00/0.00 & 0.00/1.00/0.00 \\
 & 3 & 0.138 & 1.004 & 1.003 & 0.00/0.00/1.00 & 0.00/0.00/1.00 \\
 & 4 & 0.286 & 1.002 & 1.003 & 0.00/0.00/0.99 & 0.00/0.00/1.00 \\
 & 5 & 0.558 & 1.008 & 1.004 & 0.01/0.03/0.95 & 0.00/0.00/1.00 \\
 & 6 & 0.823 & 1.017 & 1.004 & 0.05/0.13/0.82 & 0.00/0.00/1.00 \\
\bottomrule
\end{tabular}
\end{table}

\section{Per-task results on CTR23}
\label{app:pertask}
\begin{table}[h]
\centering
\caption{CTR23 ten-fold mean test RMSE per task (in the target's units). Mean: predicting the training mean.
$^{*}$: paired $t$-test $p<0.05$; $\dagger$: overlaps the development data. $N$ rows, $d$ inputs.}
\label{tab:pertask}
\scriptsize
\setlength{\tabcolsep}{3pt}
\resizebox{\textwidth}{!}{\begin{tabular}{lrrrrrrrr}
\toprule
Task & $N$ & $d$ & MJBoost & EBM & LightGBM & Mean & vs.\ EBM & vs.\ LightGBM \\
\midrule
forest fires & 517 & 12 & 48.59 & 46.88 & 46.76 & 46.66 & +3.7\% & +3.9\% \\
student performance & 649 & 30 & 2.719 & 2.776 & 2.681 & 3.191 & $-$2.1\% & +1.4\% \\
energy efficiency$^\dagger$ & 768 & 8 & 0.2969 & 0.4373 & 0.2991 & 10.08 & $-$32.1\%$^{*}$ & $-$0.7\% \\
cars & 804 & 17 & $2.12\times10^{3}$ & $2.1\times10^{3}$ & $2.17\times10^{3}$ & $9.79\times10^{3}$ & +0.9\% & $-$2.2\% \\
QSAR fish toxicity & 908 & 6 & 0.8849 & 0.8935 & 0.8839 & 1.452 & $-$1.0\% & +0.1\% \\
concrete compressive strength$^\dagger$ & 1030 & 8 & 3.974 & 3.989 & 3.851 & 16.7 & $-$0.4\% & +3.2\% \\
geographical origin of music & 1059 & 116 & 16.45 & 15.87 & 15.33 & 18.42 & +3.6\%$^{*}$ & +7.3\%$^{*}$ \\
solar flare & 1066 & 10 & 0.7714 & 0.7668 & 0.7642 & 0.8233 & +0.6\% & +0.9\% \\
socmob & 1156 & 5 & 10.85 & 14.51 & 12.94 & 40.07 & $-$25.2\%$^{*}$ & $-$16.1\%$^{*}$ \\
Moneyball & 1232 & 14 & 21.12 & 21.29 & 22.37 & 91.54 & $-$0.8\% & $-$5.6\%$^{*}$ \\
airfoil self-noise$^\dagger$ & 1503 & 5 & 1.177 & 1.903 & 1.147 & 6.895 & $-$38.1\%$^{*}$ & +2.6\% \\
red wine & 1599 & 11 & 0.5708 & 0.607 & 0.5695 & 0.807 & $-$6.0\%$^{*}$ & +0.2\% \\
auction verification & 2043 & 7 & 409.2 & $1.67\times10^{3}$ & 484.9 & $1.04\times10^{4}$ & $-$75.5\%$^{*}$ & $-$15.6\%$^{*}$ \\
space\_ga & 3107 & 6 & 0.09586 & 0.1037 & 0.1038 & 0.1974 & $-$7.6\%$^{*}$ & $-$7.6\%$^{*}$ \\
abalone$^\dagger$ & 4177 & 8 & 2.07 & 2.129 & 2.128 & 3.222 & $-$2.8\%$^{*}$ & $-$2.7\%$^{*}$ \\
white wine & 4898 & 11 & 0.5989 & 0.6583 & 0.5825 & 0.8854 & $-$9.0\%$^{*}$ & +2.8\%$^{*}$ \\
CPU activity & 8192 & 21 & 2.214 & 2.249 & 2.072 & 18.35 & $-$1.6\% & +6.8\%$^{*}$ \\
kin8nm & 8192 & 8 & 0.07093 & 0.1639 & 0.1055 & 0.2636 & $-$56.7\%$^{*}$ & $-$32.8\%$^{*}$ \\
pumadyn32nh & 8192 & 32 & 0.02384 & 0.0224 & 0.02144 & 0.03595 & +6.4\%$^{*}$ & +11.2\%$^{*}$ \\
grid stability & 10000 & 12 & $7.21\times10^{-3}$ & $8.77\times10^{-3}$ & $6.83\times10^{-3}$ & 0.03691 & $-$17.7\%$^{*}$ & +5.6\%$^{*}$ \\
Brazilian houses & 10692 & 9 & $3.77\times10^{3}$ & $9.64\times10^{3}$ & $1.18\times10^{4}$ & $1.16\times10^{4}$ & $-$60.9\% & $-$68.0\%$^{*}$ \\
naval propulsion plant & 11934 & 14 & $2.92\times10^{-5}$ & $5.07\times10^{-4}$ & $6.73\times10^{-4}$ & 0.01472 & $-$94.2\%$^{*}$ & $-$95.7\%$^{*}$ \\
Miami housing & 13932 & 15 & $8.25\times10^{4}$ & $8.82\times10^{4}$ & $8.02\times10^{4}$ & $3.17\times10^{5}$ & $-$6.5\%$^{*}$ & +2.8\%$^{*}$ \\
FIFA & 19178 & 28 & $8.93\times10^{3}$ & $8.97\times10^{3}$ & $8.92\times10^{3}$ & $1.94\times10^{4}$ & $-$0.5\% & +0.1\% \\
California housing$^\dagger$ & 20640 & 8 & $4.64\times10^{4}$ & $4.72\times10^{4}$ & $4.42\times10^{4}$ & $1.15\times10^{5}$ & $-$1.6\%$^{*}$ & +5.1\%$^{*}$ \\
superconductivity & 21263 & 81 & 9.405 & 10.04 & 9.035 & 34.25 & $-$6.3\%$^{*}$ & +4.1\%$^{*}$ \\
King County & 21613 & 21 & $1.15\times10^{5}$ & $1.2\times10^{5}$ & $1.13\times10^{5}$ & $3.66\times10^{5}$ & $-$4.2\%$^{*}$ & +1.0\% \\
health insurance & 22272 & 11 & 14.41 & 14.43 & 14.4 & 18.71 & $-$0.1\% & +0.1\%$^{*}$ \\
FPS benchmark & 24624 & 43 & 0.218 & 1.526 & 0.3636 & 54.61 & $-$85.7\%$^{*}$ & $-$40.0\%$^{*}$ \\
CPS88 wages & 28155 & 6 & 380.4 & 380.6 & 380.3 & 452 & +0.0\% & +0.0\% \\
physiochemical protein & 45730 & 9 & 3.408 & 3.968 & 3.373 & 6.118 & $-$14.1\%$^{*}$ & +1.0\%$^{*}$ \\
SARCOS & 48933 & 21 & 2.063 & 2.568 & 2.119 & 20.45 & $-$19.7\%$^{*}$ & $-$2.6\%$^{*}$ \\
diamonds & 53940 & 9 & 534.3 & 595.3 & 523.2 & $3.99\times10^{3}$ & $-$10.2\%$^{*}$ & +2.1\% \\
video transcoding & 68784 & 18 & 0.9249 & 5.599 & 0.7948 & 16.1 & $-$83.5\%$^{*}$ & +16.4\%$^{*}$ \\
wave energy & 72000 & 48 & $4.31\times10^{4}$ & $4.47\times10^{4}$ & $4.17\times10^{4}$ & $1.12\times10^{5}$ & $-$3.6\%$^{*}$ & +3.5\%$^{*}$ \\
\bottomrule
\end{tabular}}
\end{table}

\section{True Hoeffding shares of the synthetic targets}
\label{app:shares}
Every synthetic target is a product of independent one-input factors, $f=\prod_{p\le k} z_p$ with
$z_p=e^{h(x_p)/\sqrt k}$ (exponential family, $h(x)=x+\tfrac12\sin 2x$) or $z_p=x_p$ (product), and
$f=\operatorname{Im}\prod_{p\le k} w_p$ with $w_p=e^{2ih(x_p)/\sqrt k}$ (sine family). With $\mu=\E z_p$ and
$v=\Var z_p$, the part of order $j$ is $T_j=\sum_{|A|=j}\prod_{p\in A}(z_p-\mu)\prod_{p\notin A}\mu$, so
\begin{equation}
\Var T_j = \binom{k}{j} v^j \mu^{2(k-j)}
\end{equation}
for real factors, and for the sine family, with $e_p=w_p-\E w_p$,
\begin{equation}
\Var \operatorname{Im} T_j = \tfrac12\binom{k}{j}\Bigl[(\E|e|^2)^j\,|\E w|^{2(k-j)} -
\operatorname{Re}\bigl((\E e^2)^j\,(\E w)^{2(k-j)}\bigr)\Bigr].
\end{equation}
The expectations over $U(-1.5,1.5)$ are computed by 400-point Gauss--Legendre quadrature; the shares agree with
Monte Carlo estimates (e.g.\ exponential family, $k=3$: main share 0.695 against 0.694).

\section{Per-task results on CC18}
\label{app:cc18}
\begin{table}[h]
\centering
\caption{OpenML-CC18 binary tasks: ten-fold mean test log-loss (nats) and AUC per task. Prior: log-loss of the
training class frequency. \model{} vs.: ratio of the mean log-losses minus one; $^{*}$: paired $t$-test $p<0.05$.
$N$ rows, $d$ inputs.}
\label{tab:cc18pertask}
\scriptsize
\setlength{\tabcolsep}{3pt}
\resizebox{\textwidth}{!}{\begin{tabular}{lrrrrrrrrrrr}
\toprule
 & & & \multicolumn{4}{c}{Log-loss} & \multicolumn{2}{c}{\model{} vs.} & \multicolumn{3}{c}{AUC} \\
\cmidrule(lr){4-7}\cmidrule(lr){8-9}\cmidrule(lr){10-12}
Task & $N$ & $d$ & \model{} & EBM & LightGBM & Prior & EBM & LightGBM & \model{} & EBM & LightGBM \\
\midrule
dresses-sales & 500 & 12 & 0.6568 & 0.6592 & 0.6741 & 0.6803 & $-$0.4\% & $-$2.6\% & 0.6255 & 0.6348 & 0.6032 \\
kc2 & 522 & 21 & 0.3706 & 0.3806 & 0.3763 & 0.5073 & $-$2.6\% & $-$1.5\% & 0.8435 & 0.8365 & 0.8366 \\
cylinder-bands & 540 & 35 & 0.4004 & 0.3911 & 0.4037 & 0.6810 & +2.4\% & $-$0.8\% & 0.8960 & 0.9120 & 0.8923 \\
climate-model-simulation-crashes & 540 & 18 & 0.1464 & 0.1657 & 0.1424 & 0.2914 & $-$11.6\% & +2.8\% & 0.9456 & 0.9517 & 0.9413 \\
wdbc & 569 & 30 & 0.0856 & 0.1020 & 0.0981 & 0.6603 & $-$16.0\% & $-$12.7\% & 0.9963 & 0.9964 & 0.9928 \\
ilpd & 583 & 10 & 0.5065 & 0.5157 & 0.5353 & 0.5990 & $-$1.8\% & $-$5.4\%$^{*}$ & 0.7526 & 0.7360 & 0.7248 \\
credit-approval & 690 & 15 & 0.3068 & 0.3174 & 0.3023 & 0.6871 & $-$3.3\% & +1.5\% & 0.9400 & 0.9395 & 0.9444 \\
breast-w & 699 & 9 & 0.0950 & 0.0961 & 0.1026 & 0.6442 & $-$1.2\% & $-$7.4\% & 0.9939 & 0.9952 & 0.9939 \\
blood-transfusion-service-center & 748 & 4 & 0.4764 & 0.4818 & 0.4830 & 0.5487 & $-$1.1\% & $-$1.3\% & 0.7463 & 0.7484 & 0.7368 \\
diabetes & 768 & 8 & 0.4864 & 0.4719 & 0.4800 & 0.6468 & +3.1\% & +1.4\% & 0.8411 & 0.8385 & 0.8306 \\
tic-tac-toe & 958 & 9 & 0.0230 & 0.0040 & 0.0022 & 0.6453 & +473.9\%$^{*}$ & +969.7\%$^{*}$ & 0.9996 & 1.0000 & 1.0000 \\
credit-g & 1000 & 20 & 0.4895 & 0.4943 & 0.4962 & 0.6109 & $-$1.0\% & $-$1.4\% & 0.7944 & 0.7923 & 0.7883 \\
qsar-biodeg & 1055 & 41 & 0.3204 & 0.3237 & 0.3022 & 0.6393 & $-$1.0\% & +6.0\% & 0.9273 & 0.9348 & 0.9335 \\
pc1 & 1109 & 21 & 0.1948 & 0.1913 & 0.1830 & 0.2522 & +1.8\% & +6.5\% & 0.8574 & 0.8613 & 0.8756 \\
banknote-authentication & 1372 & 4 & $1.8\times10^{-9}$ & 0.0063 & 0.0094 & 0.6870 & $-$100.0\% & $-$100.0\% & 1.0000 & 1.0000 & 1.0000 \\
pc4 & 1458 & 37 & 0.2192 & 0.1894 & 0.1918 & 0.3711 & +15.7\% & +14.2\% & 0.9437 & 0.9448 & 0.9433 \\
pc3 & 1563 & 37 & 0.2729 & 0.2612 & 0.2565 & 0.3303 & +4.5\% & +6.4\% & 0.8260 & 0.8435 & 0.8447 \\
kc1 & 2109 & 21 & 0.3438 & 0.3435 & 0.3354 & 0.4306 & +0.1\% & +2.5\%$^{*}$ & 0.8235 & 0.8108 & 0.8229 \\
ozone-level-8hr & 2534 & 72 & 0.1523 & 0.1524 & 0.1508 & 0.2355 & $-$0.1\% & +1.0\% & 0.9165 & 0.9204 & 0.9155 \\
kr-vs-kp & 3196 & 36 & 0.0158 & 0.0205 & 0.0122 & 0.6922 & $-$23.0\% & +29.3\% & 0.9998 & 0.9997 & 0.9999 \\
sick & 3772 & 27 & 0.0599 & 0.0424 & 0.0262 & 0.2304 & +41.3\%$^{*}$ & +128.6\%$^{*}$ & 0.9836 & 0.9947 & 0.9984 \\
spambase & 4601 & 57 & 0.1396 & 0.1360 & 0.1201 & 0.6705 & +2.6\% & +16.2\%$^{*}$ & 0.9866 & 0.9881 & 0.9896 \\
wilt & 4839 & 5 & 0.0352 & 0.0419 & 0.0423 & 0.2099 & $-$16.0\%$^{*}$ & $-$16.8\%$^{*}$ & 0.9944 & 0.9920 & 0.9930 \\
churn & 5000 & 20 & 0.1687 & 0.1520 & 0.1448 & 0.4075 & +11.0\%$^{*}$ & +16.5\%$^{*}$ & 0.9215 & 0.9253 & 0.9209 \\
phoneme & 5404 & 5 & 0.2304 & 0.2702 & 0.2293 & 0.6052 & $-$14.7\%$^{*}$ & +0.5\% & 0.9619 & 0.9459 & 0.9612 \\
jm1 & 10885 & 21 & 0.4383 & 0.4305 & 0.4213 & 0.4912 & +1.8\%$^{*}$ & +4.0\%$^{*}$ & 0.7290 & 0.7418 & 0.7582 \\
PhishingWebsites & 11055 & 30 & 0.0712 & 0.1002 & 0.0609 & 0.6866 & $-$29.0\%$^{*}$ & +16.8\%$^{*}$ & 0.9968 & 0.9937 & 0.9976 \\
nomao & 34465 & 118 & 0.0841 & 0.0814 & 0.0673 & 0.5982 & +3.4\%$^{*}$ & +24.9\%$^{*}$ & 0.9946 & 0.9949 & 0.9965 \\
bank-marketing & 45211 & 16 & 0.2126 & 0.1998 & 0.1909 & 0.3609 & +6.4\%$^{*}$ & +11.4\%$^{*}$ & 0.9306 & 0.9340 & 0.9392 \\
electricity & 45312 & 8 & 0.2783 & 0.2529 & 0.1432 & 0.6817 & +10.1\%$^{*}$ & +94.3\%$^{*}$ & 0.9568 & 0.9607 & 0.9870 \\
adult & 48842 & 14 & 0.2802 & 0.2731 & 0.2732 & 0.5503 & +2.6\%$^{*}$ & +2.5\%$^{*}$ & 0.9284 & 0.9301 & 0.9304 \\
numerai28.6 & 96320 & 21 & 0.6917 & 0.6917 & 0.6918 & 0.6931 & +0.0\% & +0.0\% & 0.5298 & 0.5297 & 0.5280 \\
\bottomrule
\end{tabular}}
\end{table}

\begin{table}[h]
\centering
\caption{OpenML-CC18 multiclass tasks: ten-fold mean test log-loss (nats) and accuracy per task. Prior: log-loss of
the training class frequencies. \model{} vs.: ratio of the mean log-losses minus one; $^{*}$: paired $t$-test
$p<0.05$. $N$ rows, $d$ inputs, $K$ classes.}
\label{tab:mcpertask}
\scriptsize
\setlength{\tabcolsep}{3pt}
\resizebox{\textwidth}{!}{\begin{tabular}{lrrrrrrrrrrrr}
\toprule
 & & & & \multicolumn{4}{c}{Log-loss} & \multicolumn{2}{c}{\model{} vs.} & \multicolumn{3}{c}{Accuracy} \\
\cmidrule(lr){5-8}\cmidrule(lr){9-10}\cmidrule(lr){11-13}
Task & $N$ & $d$ & $K$ & \model{} & EBM & LightGBM & Prior & EBM & LightGBM & \model{} & EBM & LightGBM \\
\midrule
balance-scale & 625 & 4 & 3 & 0.0122 & 0.0450 & 0.2981 & 0.9137 & $-$73.0\%$^{*}$ & $-$95.9\%$^{*}$ & 0.9952 & 0.9840 & 0.8880 \\
eucalyptus & 736 & 19 & 5 & 0.7517 & 0.7210 & 0.7457 & 1.5680 & +4.3\%$^{*}$ & +0.8\% & 0.6725 & 0.6819 & 0.6941 \\
analcatdata\_dmft & 797 & 4 & 6 & 1.7726 & 1.7442 & 1.7622 & 1.7886 & +1.6\%$^{*}$ & +0.6\% & 0.2020 & 0.2159 & 0.1983 \\
vehicle & 846 & 18 & 4 & 0.3806 & 0.4298 & 0.4581 & 1.3857 & $-$11.5\%$^{*}$ & $-$16.9\%$^{*}$ & 0.8060 & 0.7860 & 0.7577 \\
vowel & 990 & 12 & 11 & 0.0851 & 0.1088 & 0.1390 & 2.3979 & $-$21.8\% & $-$38.8\%$^{*}$ & 0.9737 & 0.9606 & 0.9616 \\
cmc & 1473 & 9 & 3 & 0.8857 & 0.8948 & 0.9009 & 1.0668 & $-$1.0\% & $-$1.7\%$^{*}$ & 0.5642 & 0.5642 & 0.5384 \\
car & 1728 & 6 & 4 & 0.0020 & 0.0874 & 0.0092 & 0.8358 & $-$97.7\%$^{*}$ & $-$77.8\%$^{*}$ & 0.9994 & 0.9664 & 0.9983 \\
steel-plates-fault & 1941 & 27 & 7 & 0.5428 & 0.5717 & 0.4853 & 1.6710 & $-$5.1\% & +11.9\%$^{*}$ & 0.7877 & 0.7826 & 0.8042 \\
mfeat-morphological & 2000 & 6 & 10 & 0.6117 & 0.6201 & 0.6632 & 2.3026 & $-$1.4\% & $-$7.8\%$^{*}$ & 0.7375 & 0.7290 & 0.7150 \\
segment & 2310 & 16 & 7 & 0.1515 & 0.1871 & 0.1721 & 1.9459 & $-$19.0\%$^{*}$ & $-$11.9\%$^{*}$ & 0.9450 & 0.9338 & 0.9381 \\
wall-robot-navigation & 5456 & 24 & 4 & 0.0178 & 0.0063 & 0.0141 & 1.1885 & +183.6\%$^{*}$ & +26.2\% & 0.9923 & 0.9985 & 0.9956 \\
\bottomrule
\end{tabular}}
\end{table}


\begin{thebibliography}{42}
\providecommand{\natexlab}[1]{#1}
\providecommand{\url}[1]{\texttt{#1}}
\expandafter\ifx\csname urlstyle\endcsname\relax
  \providecommand{\doi}[1]{doi: #1}\else
  \providecommand{\doi}{doi: \begingroup \urlstyle{rm}\Url}\fi

\bibitem[Agarwal et~al.(2021)Agarwal, Melnick, Frosst, Zhang, Lengerich,
  Caruana, and Hinton]{agarwal2021nam}
Rishabh Agarwal, Levi Melnick, Nicholas Frosst, Xuezhou Zhang, Ben Lengerich,
  Rich Caruana, and Geoffrey~E. Hinton.
\newblock Neural additive models: Interpretable machine learning with neural
  nets.
\newblock In \emph{Advances in Neural Information Processing Systems},
  volume~34, pages 4699--4711, 2021.
\newblock URL
  \url{https://papers.nips.cc/paper_files/paper/2021/hash/251bd0442dfcc53b5a761e050f8022b8-Abstract.html}.

\bibitem[Bischl et~al.(2021)Bischl, Casalicchio, Feurer, Gijsbers, Hutter,
  Lang, Gomes~Mantovani, van Rijn, and Vanschoren]{bischl2021openml}
Bernd Bischl, Giuseppe Casalicchio, Matthias Feurer, Pieter Gijsbers, Frank
  Hutter, Michel Lang, Rafael Gomes~Mantovani, Jan van Rijn, and Joaquin
  Vanschoren.
\newblock {OpenML} benchmarking suites.
\newblock In \emph{Proceedings of the Neural Information Processing Systems
  Track on Datasets and Benchmarks}, volume~1, 2021.
\newblock URL
  \url{https://datasets-benchmarks-proceedings.neurips.cc/paper_files/paper/2021/hash/c7e1249ffc03eb9ded908c236bd1996d-Abstract-round2.html}.

\bibitem[Breiman and Friedman(1985)]{breiman1985ace}
Leo Breiman and Jerome~H. Friedman.
\newblock Estimating optimal transformations for multiple regression and
  correlation.
\newblock \emph{Journal of the American Statistical Association}, 80\penalty0
  (391):\penalty0 580--598, 1985.
\newblock \doi{10.1080/01621459.1985.10478157}.

\bibitem[B{\"u}hlmann and Hothorn(2007)]{buhlmann2007boosting}
Peter B{\"u}hlmann and Torsten Hothorn.
\newblock Boosting algorithms: Regularization, prediction and model fitting.
\newblock \emph{Statistical Science}, 22\penalty0 (4):\penalty0 477--505, 2007.
\newblock \doi{10.1214/07-sts242}.

\bibitem[Caruana et~al.(2015)Caruana, Lou, Gehrke, Koch, Sturm, and
  Elhadad]{caruana2015intelligible}
Rich Caruana, Yin Lou, Johannes Gehrke, Paul Koch, Marc Sturm, and Noemie
  Elhadad.
\newblock Intelligible models for {HealthCare}: Predicting pneumonia risk and
  hospital 30-day readmission.
\newblock In \emph{Proceedings of the 21st ACM SIGKDD International Conference
  on Knowledge Discovery and Data Mining}, pages 1721--1730. ACM, 2015.
\newblock \doi{10.1145/2783258.2788613}.

\bibitem[Chang et~al.(2022)Chang, Caruana, and Goldenberg]{chang2022nodegam}
Chun-Hao Chang, Rich Caruana, and Anna Goldenberg.
\newblock {NODE-GAM}: Neural generalized additive model for interpretable deep
  learning.
\newblock In \emph{International Conference on Learning Representations}, 2022.
\newblock URL \url{https://iclr.cc/virtual/2022/poster/6355}.

\bibitem[Chen and Guestrin(2016)]{chen2016xgboost}
Tianqi Chen and Carlos Guestrin.
\newblock {XGBoost}: A scalable tree boosting system.
\newblock In \emph{Proceedings of the 22nd ACM SIGKDD International Conference
  on Knowledge Discovery and Data Mining}, pages 785--794. ACM, 2016.
\newblock \doi{10.1145/2939672.2939785}.

\bibitem[Coppejans(2004)]{coppejans2004kolmogorov}
Mark Coppejans.
\newblock On {Kolmogorov's} representation of functions of several variables by
  functions of one variable.
\newblock \emph{Journal of Econometrics}, 123\penalty0 (1):\penalty0 1--31,
  2004.
\newblock \doi{10.1016/j.jeconom.2003.10.026}.

\bibitem[Eilers and Marx(1996)]{eilers1996flexible}
Paul H.~C. Eilers and Brian~D. Marx.
\newblock Flexible smoothing with {B-splines} and penalties.
\newblock \emph{Statistical Science}, 11\penalty0 (2):\penalty0 89--121, 1996.
\newblock \doi{10.1214/ss/1038425655}.

\bibitem[Fischer et~al.(2023)Fischer, Feurer, and Bischl]{fischer2023ctr23}
Sebastian~Felix Fischer, Matthias Feurer, and Bernd Bischl.
\newblock {OpenML-CTR23} -- a curated tabular regression benchmarking suite.
\newblock In \emph{AutoML 2023 (Workshop)}, 2023.
\newblock URL \url{https://openreview.net/forum?id=HebAOoMm94}.

\bibitem[Friedman et~al.(2000)Friedman, Hastie, and
  Tibshirani]{friedman2000additive}
Jerome Friedman, Trevor Hastie, and Robert Tibshirani.
\newblock Additive logistic regression: a statistical view of boosting (with
  discussion and a rejoinder by the authors).
\newblock \emph{The Annals of Statistics}, 28\penalty0 (2):\penalty0 337--407,
  2000.
\newblock \doi{10.1214/aos/1016218223}.

\bibitem[Friedman(2001)]{friedman2001greedy}
Jerome~H. Friedman.
\newblock Greedy function approximation: A gradient boosting machine.
\newblock \emph{The Annals of Statistics}, 29\penalty0 (5):\penalty0
  1189--1232, 2001.
\newblock \doi{10.1214/aos/1013203451}.

\bibitem[Friedman and Stuetzle(1981)]{friedman1981ppr}
Jerome~H. Friedman and Werner Stuetzle.
\newblock Projection pursuit regression.
\newblock \emph{Journal of the American Statistical Association}, 76\penalty0
  (376):\penalty0 817--823, 1981.
\newblock \doi{10.1080/01621459.1981.10477729}.

\bibitem[Grinsztajn et~al.(2022)Grinsztajn, Oyallon, and
  Varoquaux]{grinsztajn2022tree}
Leo Grinsztajn, Edouard Oyallon, and Gael Varoquaux.
\newblock Why do tree-based models still outperform deep learning on typical
  tabular data?
\newblock In \emph{Advances in Neural Information Processing Systems},
  volume~35, pages 507--520, 2022.
\newblock \doi{10.52202/068431-0037}.
\newblock URL
  \url{https://papers.nips.cc/paper_files/paper/2022/hash/0378c7692da36807bdec87ab043cdadc-Abstract-Datasets_and_Benchmarks.html}.

\bibitem[Hastie and Tibshirani(1986)]{hastie1986gam}
Trevor Hastie and Robert Tibshirani.
\newblock Generalized additive models.
\newblock \emph{Statistical Science}, 1\penalty0 (3):\penalty0 297--310, 1986.
\newblock \doi{10.1214/ss/1177013604}.

\bibitem[Hoeffding(1948)]{hoeffding1948class}
Wassily Hoeffding.
\newblock A class of statistics with asymptotically normal distribution.
\newblock \emph{The Annals of Mathematical Statistics}, 19\penalty0
  (3):\penalty0 293--325, 1948.
\newblock \doi{10.1214/aoms/1177730196}.

\bibitem[Hooker(2007)]{hooker2007generalized}
Giles Hooker.
\newblock Generalized functional {ANOVA} diagnostics for high-dimensional
  functions of dependent variables.
\newblock \emph{Journal of Computational and Graphical Statistics}, 16\penalty0
  (3):\penalty0 709--732, 2007.
\newblock \doi{10.1198/106186007x237892}.

\bibitem[Horowitz and Mammen(2007)]{horowitz2007rate}
Joel~L. Horowitz and Enno Mammen.
\newblock Rate-optimal estimation for a general class of nonparametric
  regression models with unknown link functions.
\newblock \emph{The Annals of Statistics}, 35\penalty0 (6):\penalty0
  2589--2619, 2007.
\newblock \doi{10.1214/009053607000000415}.

\bibitem[Hwang et~al.(1994)Hwang, Lay, Maechler, Martin, and
  Schimert]{hwang1994regression}
Jeng-Neng Hwang, Shyh-Rong Lay, M.~Maechler, R.~D. Martin, and J.~Schimert.
\newblock Regression modeling in back-propagation and projection pursuit
  learning.
\newblock \emph{IEEE Transactions on Neural Networks}, 5\penalty0 (3):\penalty0
  342--353, 1994.
\newblock \doi{10.1109/72.286906}.

\bibitem[Ke et~al.(2017)Ke, Meng, Finley, Wang, Chen, Ma, Ye, and
  Liu]{ke2017lightgbm}
Guolin Ke, Qi~Meng, Thomas Finley, Taifeng Wang, Wei Chen, Weidong Ma, Qiwei
  Ye, and Tie-Yan Liu.
\newblock {LightGBM}: A highly efficient gradient boosting decision tree.
\newblock In \emph{Advances in Neural Information Processing Systems},
  volume~30, 2017.
\newblock URL
  \url{https://papers.nips.cc/paper_files/paper/2017/hash/6449f44a102fde848669bdd9eb6b76fa-Abstract.html}.

\bibitem[Kolodyazhniy and Bodyanskiy(2004)]{kolodyazhniy2004fuzzy}
Vitaliy Kolodyazhniy and Yevgeniy Bodyanskiy.
\newblock Fuzzy {Kolmogorov's} network.
\newblock In \emph{Knowledge-Based Intelligent Information and Engineering
  Systems (KES 2004), Lecture Notes in Computer Science 3214}, pages 764--771.
  Springer Berlin Heidelberg, 2004.
\newblock ISBN 9783540301332.
\newblock \doi{10.1007/978-3-540-30133-2_100}.

\bibitem[Lam et~al.(2015)Lam, Pitrou, and Seibert]{lam2015numba}
Siu~Kwan Lam, Antoine Pitrou, and Stanley Seibert.
\newblock {Numba}: a {LLVM-based} {Python} {JIT} compiler.
\newblock In \emph{Proceedings of the Second Workshop on the LLVM Compiler
  Infrastructure in HPC}, pages 1--6. ACM, 2015.
\newblock \doi{10.1145/2833157.2833162}.

\bibitem[Lengerich et~al.(2020)Lengerich, Tan, Chang, Hooker, and
  Caruana]{lengerich2020purifying}
Benjamin Lengerich, Sarah Tan, Chun-Hao Chang, Giles Hooker, and Rich Caruana.
\newblock Purifying interaction effects with the functional {ANOVA}: An
  efficient algorithm for recovering identifiable additive models.
\newblock In \emph{Proceedings of the International Conference on Artificial
  Intelligence and Statistics (PMLR)}, volume 108, pages 2402--2412, 2020.
\newblock URL \url{https://proceedings.mlr.press/v108/lengerich20a.html}.

\bibitem[Li(1992)]{li1992phd}
Ker-Chau Li.
\newblock On principal {Hessian} directions for data visualization and
  dimension reduction: Another application of {Stein's} lemma.
\newblock \emph{Journal of the American Statistical Association}, 87\penalty0
  (420):\penalty0 1025--1039, 1992.
\newblock \doi{10.1080/01621459.1992.10476258}.

\bibitem[Lin et~al.(2018)Lin, Pan, Lv, and Zhang]{lin2018efficient}
Huazhen Lin, Lixian Pan, Shaogao Lv, and Wenyang Zhang.
\newblock Efficient estimation and computation for the generalised additive
  models with unknown link function.
\newblock \emph{Journal of Econometrics}, 202\penalty0 (2):\penalty0 230--244,
  2018.
\newblock \doi{10.1016/j.jeconom.2017.11.001}.

\bibitem[Liu et~al.(2025)Liu, Wang, Vaidya, Ruehle, Halverson,
  Solja{\v{c}}i{\'c}, Hou, and Tegmark]{liu2024kan}
Ziming Liu, Yixuan Wang, Sachin Vaidya, Fabian Ruehle, James Halverson, Marin
  Solja{\v{c}}i{\'c}, Thomas~Y. Hou, and Max Tegmark.
\newblock {KAN}: {Kolmogorov--Arnold} networks.
\newblock In \emph{International Conference on Learning Representations}, 2025.
\newblock URL \url{https://iclr.cc/virtual/2025/oral/31858}.

\bibitem[Lou et~al.(2012)Lou, Caruana, and Gehrke]{lou2012intelligible}
Yin Lou, Rich Caruana, and Johannes Gehrke.
\newblock Intelligible models for classification and regression.
\newblock In \emph{Proceedings of the 18th ACM SIGKDD international conference
  on Knowledge discovery and data mining}, pages 150--158. ACM, 2012.
\newblock \doi{10.1145/2339530.2339556}.

\bibitem[Lou et~al.(2013)Lou, Caruana, Gehrke, and Hooker]{lou2013accurate}
Yin Lou, Rich Caruana, Johannes Gehrke, and Giles Hooker.
\newblock Accurate intelligible models with pairwise interactions.
\newblock In \emph{Proceedings of the 19th ACM SIGKDD international conference
  on Knowledge discovery and data mining}, pages 623--631. ACM, 2013.
\newblock \doi{10.1145/2487575.2487579}.

\bibitem[Lundberg and Lee(2017)]{lundberg2017shap}
Scott~M. Lundberg and Su-In Lee.
\newblock A unified approach to interpreting model predictions.
\newblock In \emph{Advances in Neural Information Processing Systems},
  volume~30, 2017.
\newblock URL
  \url{https://papers.nips.cc/paper_files/paper/2017/hash/8a20a8621978632d76c43dfd28b67767-Abstract.html}.

\bibitem[Micci-Barreca(2001)]{micci2001preprocessing}
Daniele Micci-Barreca.
\newblock A preprocessing scheme for high-cardinality categorical attributes in
  classification and prediction problems.
\newblock \emph{ACM SIGKDD Explorations Newsletter}, 3\penalty0 (1):\penalty0
  27--32, 2001.
\newblock \doi{10.1145/507533.507538}.

\bibitem[Mohr and Frochte(2026)]{mohr2026gbkan}
Janis Mohr and J{\"o}rg Frochte.
\newblock {GB-KAN}: Gradient boosting with interpretable {Kolmogorov-Arnold}
  networks.
\newblock In \emph{Proceedings of the 18th International Conference on Agents
  and Artificial Intelligence}, pages 2421--2432. SCITEPRESS - Science and
  Technology Publications, 2026.
\newblock \doi{10.5220/0014246800004052}.

\bibitem[Nori et~al.(2019)Nori, Jenkins, Koch, and
  Caruana]{nori2019interpretml}
Harsha Nori, Samuel Jenkins, Paul Koch, and Rich Caruana.
\newblock {InterpretML}: A unified framework for machine learning
  interpretability.
\newblock arXiv:1909.09223, 2019.

\bibitem[Pednault(2006)]{pednault2006transform}
Edwin Pednault.
\newblock Transform regression and the {Kolmogorov} superposition theorem.
\newblock In \emph{Proceedings of the 2006 SIAM International Conference on
  Data Mining}, pages 35--46. Society for Industrial and Applied Mathematics,
  2006.
\newblock \doi{10.1137/1.9781611972764.4}.

\bibitem[Prokhorenkova et~al.(2018)Prokhorenkova, Gusev, Vorobev, Dorogush, and
  Gulin]{prokhorenkova2018catboost}
Liudmila Prokhorenkova, Gleb Gusev, Aleksandr Vorobev, Anna~Veronika Dorogush,
  and Andrey Gulin.
\newblock {CatBoost}: unbiased boosting with categorical features.
\newblock In \emph{Advances in Neural Information Processing Systems},
  volume~31, 2018.
\newblock URL
  \url{https://papers.nips.cc/paper_files/paper/2018/hash/14491b756b3a51daac41c24863285549-Abstract.html}.

\bibitem[Rudin(2019)]{rudin2019stop}
Cynthia Rudin.
\newblock Stop explaining black box machine learning models for high stakes
  decisions and use interpretable models instead.
\newblock \emph{Nature Machine Intelligence}, 1\penalty0 (5):\penalty0
  206--215, 2019.
\newblock \doi{10.1038/s42256-019-0048-x}.

\bibitem[Schmid and Hothorn(2008)]{schmid2008boosting}
Matthias Schmid and Torsten Hothorn.
\newblock Boosting additive models using component-wise {P-Splines}.
\newblock \emph{Computational Statistics \& Data Analysis}, 53\penalty0
  (2):\penalty0 298--311, 2008.
\newblock \doi{10.1016/j.csda.2008.09.009}.

\bibitem[Sobol'(2001)]{sobol2001global}
I.~M. Sobol'.
\newblock Global sensitivity indices for nonlinear mathematical models and
  their {Monte} {Carlo} estimates.
\newblock \emph{Mathematics and Computers in Simulation}, 55\penalty0
  (1-3):\penalty0 271--280, 2001.
\newblock \doi{10.1016/s0378-4754(00)00270-6}.

\bibitem[Spiegel et~al.(2019)Spiegel, Kneib, and
  Otto-Sobotka]{spiegel2019flexible}
Elmar Spiegel, Thomas Kneib, and Fabian Otto-Sobotka.
\newblock Generalized additive models with flexible response functions.
\newblock \emph{Statistics and Computing}, 29\penalty0 (1):\penalty0 123--138,
  2019.
\newblock \doi{10.1007/s11222-017-9799-6}.

\bibitem[Tutz and Petry(2016)]{tutz2016generalized}
Gerhard Tutz and Sebastian Petry.
\newblock Generalized additive models with unknown link function including
  variable selection.
\newblock \emph{Journal of Applied Statistics}, 43\penalty0 (15):\penalty0
  2866--2885, 2016.
\newblock \doi{10.1080/02664763.2016.1155109}.

\bibitem[Vanschoren et~al.(2014)Vanschoren, van Rijn, Bischl, and
  Torgo]{vanschoren2014openml}
Joaquin Vanschoren, Jan~N. van Rijn, Bernd Bischl, and Luis Torgo.
\newblock {OpenML}: networked science in machine learning.
\newblock \emph{ACM SIGKDD Explorations Newsletter}, 15\penalty0 (2):\penalty0
  49--60, 2014.
\newblock \doi{10.1145/2641190.2641198}.

\bibitem[Wood and Fasiolo(2017)]{wood2017generalized}
Simon~N. Wood and Matteo Fasiolo.
\newblock A generalized {Fellner-Schall} method for smoothing parameter
  optimization with application to {Tweedie} location, scale and shape models.
\newblock \emph{Biometrics}, 73\penalty0 (4):\penalty0 1071--1081, 2017.
\newblock \doi{10.1111/biom.12666}.

\bibitem[Yang et~al.(2021)Yang, Zhang, and Sudjianto]{yang2021gaminet}
Zebin Yang, Aijun Zhang, and Agus Sudjianto.
\newblock {GAMI-Net}: An explainable neural network based on generalized
  additive models with structured interactions.
\newblock \emph{Pattern Recognition}, 120:\penalty0 108192, 2021.
\newblock \doi{10.1016/j.patcog.2021.108192}.

\end{thebibliography}
\end{document}